\documentclass{article}

\usepackage{PRIMEarxiv}

\usepackage[utf8]{inputenc} 
\usepackage[T1]{fontenc}    
\usepackage{hyperref}       
\usepackage{url}            
\usepackage{booktabs}       
\usepackage{amsfonts}       
\usepackage{nicefrac}       
\usepackage{microtype}      
\usepackage{lipsum}
\usepackage{fancyhdr}       
\usepackage{graphicx}       

\usepackage{balance}
\usepackage{multirow}
\usepackage{caption}
\usepackage{subcaption}
\usepackage{footnote}   
\usepackage[misc]{ifsym}
\usepackage{color, colortbl}
 \usepackage{amsmath}
\usepackage[figuresright]{rotating}

\definecolor{Gray}{gray}{0.9}
\definecolor{LightCyan}{rgb}{0.88,1,1}
\newtheorem{myDef}{Definition}

\graphicspath{{media/}}     

\title{LightPath: Lightweight and Scalable Path Representation Learning
}
\author{
  Sean Bin Yang \\
  Aalborg University, Denmark\\
  \texttt{sean.byang@gmail.com} \\
   \And
  Jilin Hu \thanks{Corresponding Author: Jilin Hu (jlhu@dase.ecnu.edu.cn)} \\
  East China Normal University, China\\
  Aalborg University, Denmark \\
  \texttt{jlhu@dase.ecnu.edu.cn} \\
  \And
  Chenjuan Guo  \\
  East China Normal University, China\\
  Aalborg University, Denmark \\
  \texttt{cjguo@dase.ecnu.edu.cn} \\
  \And
  Bin Yang  \\
  East China Normal University, China\\
  Aalborg University, Denmark \\
  \texttt{byang@dase.ecnu.edu.cn} \\
  \And
  Christian S. Jensen\\
  Aalborg University, Denmark\\
  \texttt{csj@cs.aau.dk} \\
}

\begin{document}
\maketitle

\begin{abstract}
Movement paths are used widely in intelligent transportation and smart city applications. To serve such applications, path representation learning aims to provide compact representations of paths that enable efficient and accurate operations when used for different downstream tasks such as path ranking and travel cost estimation. In many cases, it is attractive that the path representation learning is lightweight and scalable; in resource-limited environments and under green computing limitations, it is essential. Yet, existing path representation learning studies focus on accuracy and pay at most secondary attention to resource consumption and scalability.
		We propose a lightweight and scalable path representation learning framework, termed \textbf{\emph{LightPath}}, that aims to reduce resource consumption and achieve scalability without affecting accuracy, thus enabling broader applicability. More specifically, we first propose a sparse auto-encoder that ensures that the framework achieves good scalability with respect to path length. Next, we propose a relational reasoning framework to enable faster training of more robust sparse path encoders. We also propose global-local knowledge distillation to further reduce the size and improve the performance of sparse path encoders. Finally, we report extensive experiments on two real-world datasets to offer insight into the efficiency, scalability, and effectiveness of the proposed framework.
\end{abstract}

\keywords{Path representation learning 
\and Lightweight \and Self-supervised learning}

\section{Introduction}

\label{sec:intro}
\begin{figure}[t]
    \centering
    \includegraphics[scale=0.6]{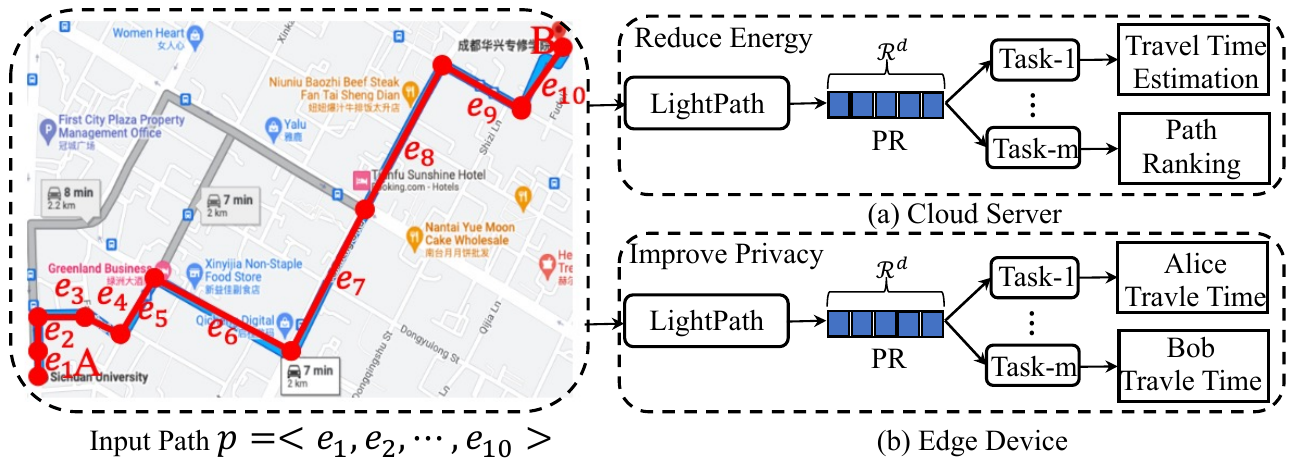}
    \caption{Intuition of the Lightweight Path Representation Learning Problem}
    \label{fig:re}

\end{figure}

Motivated in part by an increasing number of intelligent transportation and smart city services that operate on movement paths, path representation learning (PRL) has received remarkable attention~\cite{DBLP:conf/icde/Guo0HJ18,DBLP:journals/pvldb/FangPCDG21,DBLP:journals/tkde/ZhangZSQ20}.
Path representation learning aims to learn a generic path representation (PR) vector (ref. $\mathcal{R}^d$ in Figure \ref{fig:re}) that can be utilized in a range of different downstream tasks. This is in contrast to task-specific path representation learning performed by supervised methods that yield representations that work well on task-labeled data but work poorly in other tasks. 
For example, in Figure \ref{fig:re} \emph{Lightpath} takes as input a path $p$ and returns a generic PR that can support a variety of tasks, e.g., travel time estimation and path ranking.

In fact, a variety of intelligent transportation services involve paths, e.g., travel cost estimation~\cite{DBLP:conf/ijcai/YangGHT021,DBLP:conf/icde/HuG0J19,DBLP:conf/sigmod/Yuan0BF20,DBLP:journals/pvldb/WuZGHYJ21,DBLP:conf/icde/00020BF21,DBLP:journals/pvldb/TranMLYHS20}, 
%
trajectory analysis~\cite{DBLP:conf/sigmod/Shang0B18,DBLP:conf/sigmod/Shang0B18a,DBLP:journals/pvldb/HanCMG22,DBLP:journals/pvldb/WangLCL20,DBLP:journals/pvldb/WangBCSQ19,DBLP:journals/vldb/PedersenYJ20,DBLP:journals/sigmod/GuoJ014,DBLP:journals/pvldb/PedersenYJ20}, and path ranking~\cite{yang2020context,DBLP:journals/pvldb/0026HFZL020,DBLP:conf/ijcai/YangGHT021,DBLP:journals/vldb/GuoYHJC20,DBLP:conf/icde/LiuJYZ18}. Path representations that are both accurate and compact, thus facilitating efficient operations, are in high demand as they hold the potential to significantly improve the services that use them.
Indeed, 
recent path representation learning methods, in particular deep learning based methods, demonstrate impressive and state-of-the-art performance on a wide variety of downstream tasks. 

However, existing path representation learning methods focus on accuracy improvement and pay at best secondary attention to scalability and resource usage. The resulting models often include large numbers of layers and parameters, driving up computational costs, power consumption, and memory consumption, especially for long paths.
%
Although path encoders with many parameters may achieve good accuracy, they have two limitations. First, using large path encoders in the cloud consumes substantial energy, which is not eco-friendly (cf. Fig 1(a)). Second, increasingly many users enjoy personalized services, e.g., personalized travel time estimation based on their own trajectories. Due to privacy concerns, such personalized services often require the path encoder to be deployed in resource-limited edge environments, such as on individual users' mobile phones (cf. Fig 1(b)), without having to upload their trajectories to the cloud.
More generally, it is sensible to enable lightweight path representation learning that works in resource-limited environments.

Next, existing path representation methods suffer from two limitations. 

%

\begin{figure}[b]
     \centering
     \begin{subfigure}[b]{0.45\textwidth}
         \centering
         \includegraphics[width=\textwidth]{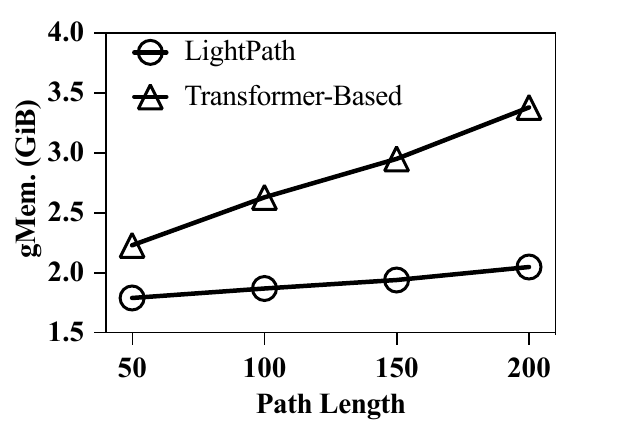}
         \caption{GPU Memory}
         \label{fig:subfig:mac}
     \end{subfigure}
     \hfill
     \begin{subfigure}[b]{0.45\textwidth}
         \centering
         \includegraphics[width=\textwidth]{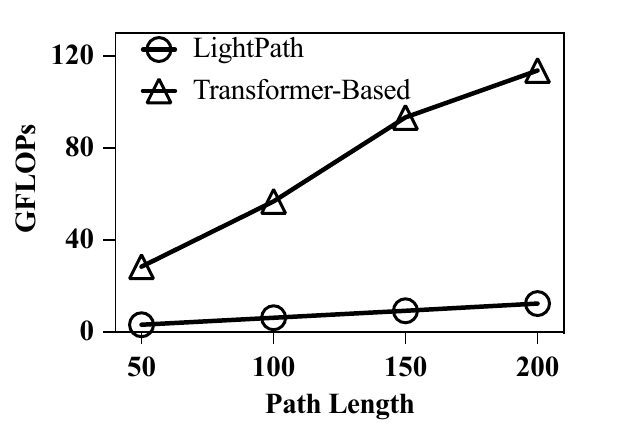}
         \caption{GFLOPs}
         \label{fig:subfig:flops}
     \end{subfigure}
        \caption{Scalability w.r.t. Path Length.}
        \label{fig:scale}

\end{figure}
\begin{table*}[t]
\centering
\caption{Model Parameter Size with Varying Encoder Layers}

\begin{tabular}{l|l|l|l|l}
\toprule[2pt]
Encoder Layers L                                                  & 12     & 24      & 48     & 96      \\ \toprule[1pt]
\begin{tabular}[c]{@{}l@{}}Parameters\\ (Millions)\end{tabular} & 29.85 & 55.07 & 105.51 & 206.40 \\ \toprule[2pt]
\end{tabular}
\label{tab:mp}

\end{table*}

\begin{figure*}[t]
     \centering
     \begin{subfigure}[b]{0.3\textwidth}
         \centering
         \includegraphics[scale=0.31]{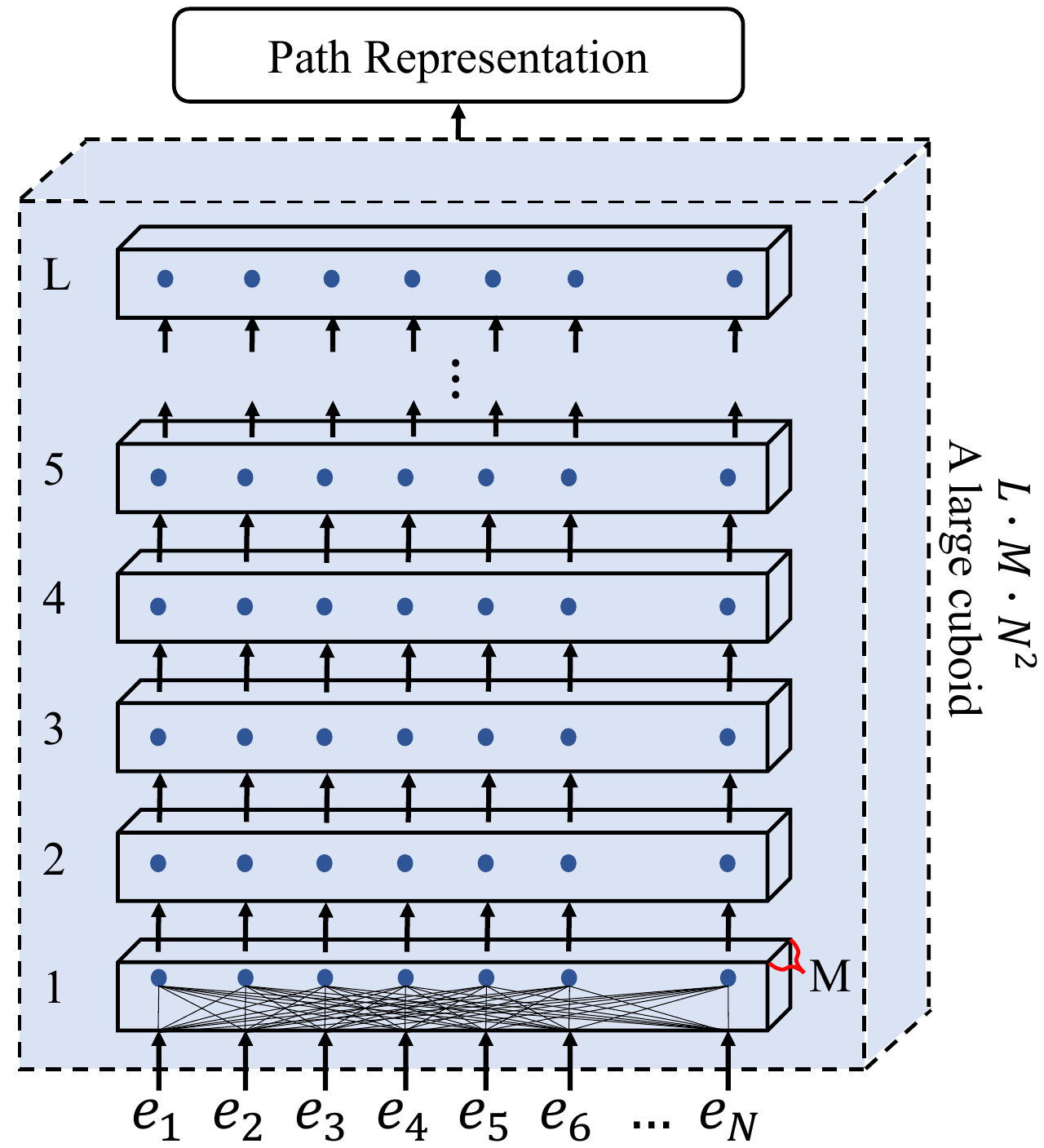}
         \caption{Traditional Transformer Encoder}
         \label{fig:subfig:tte}
     \end{subfigure}
     \hfill
     \begin{subfigure}[b]{0.3\textwidth}
         \centering
         \includegraphics[scale=0.31]{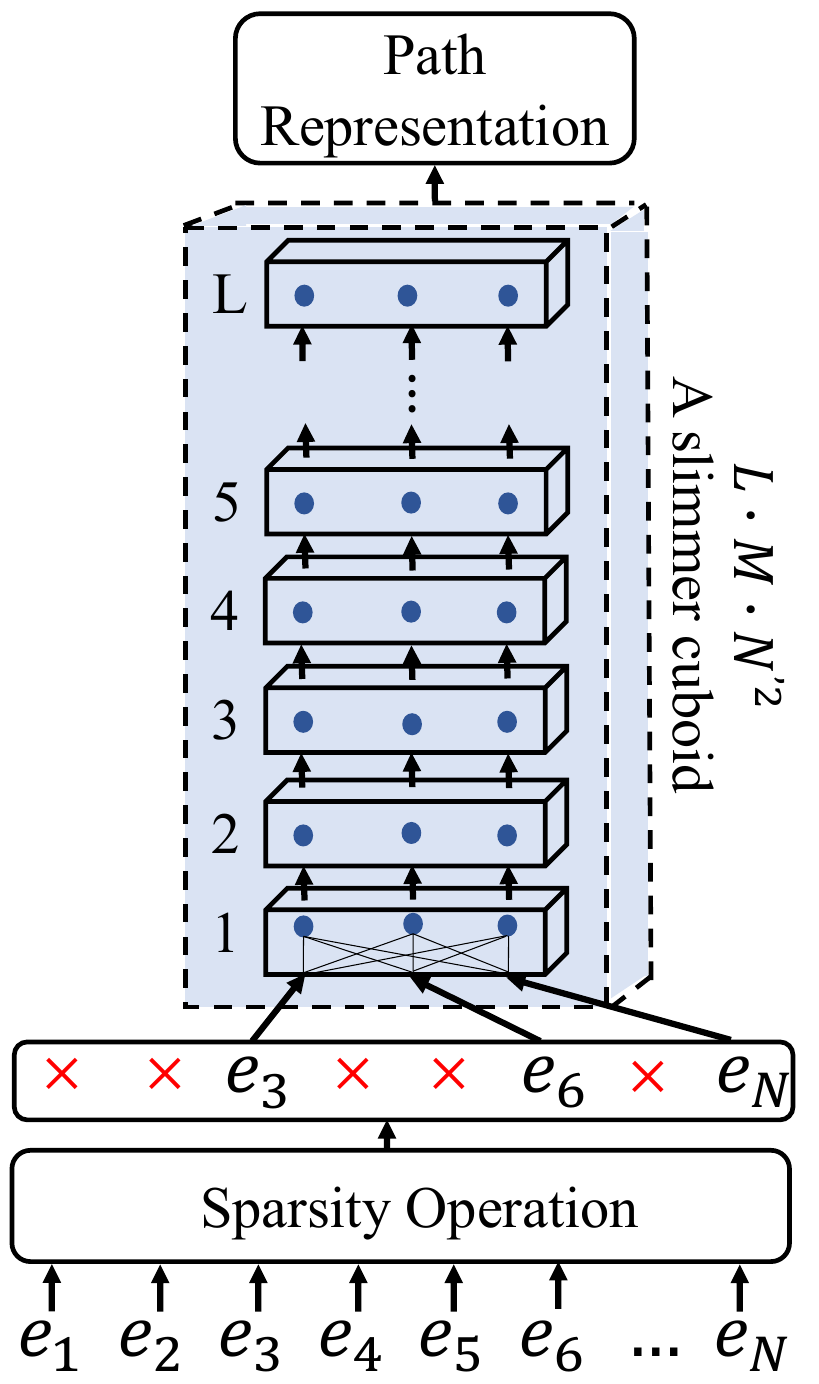}
         \caption{Sparse Transformer Encoder}
         \label{fig:subfig:ste}
     \end{subfigure}
     \begin{subfigure}[b]{0.3\textwidth}
         \centering
    \includegraphics[scale=0.31]{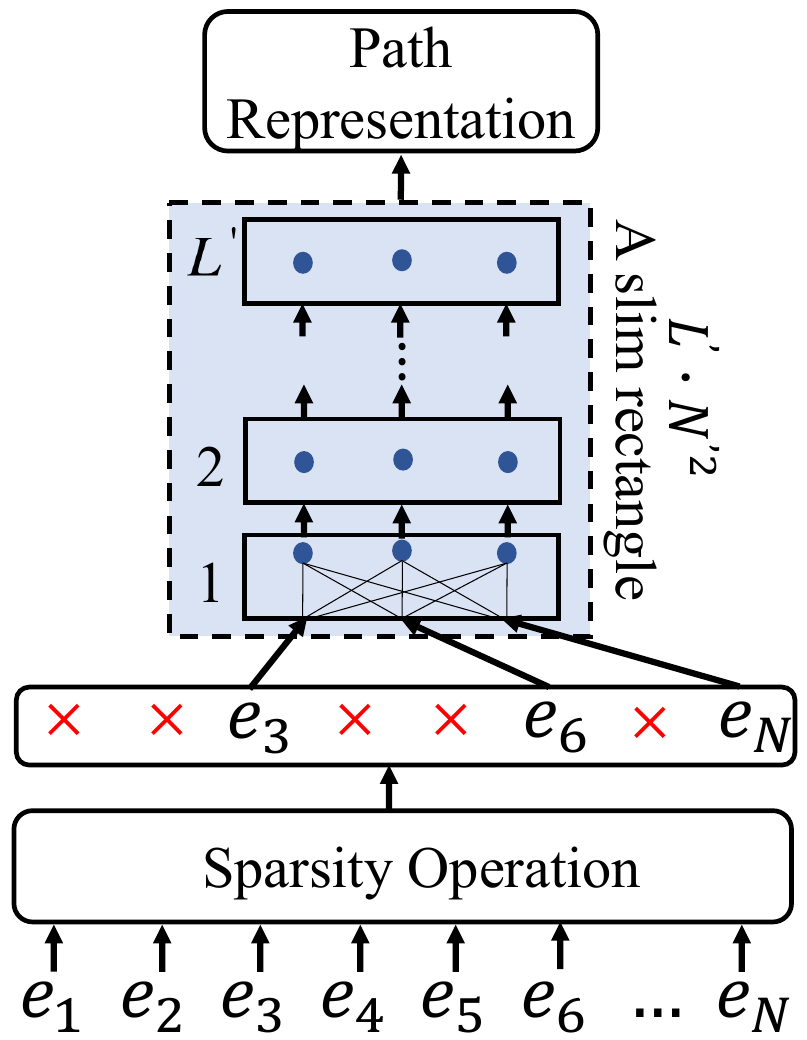}
    \caption{LightPath}
    \label{fig:subfig:lp}
     \end{subfigure}
        \caption{Encoder Architectures: (a) A traditional transformer encoder with $L$ layers and $M$ heads, takes as input a path ($N$-Length) and has complexity $\mathcal{O}\left(L\cdot M \cdot N^{2}\right)$; (b) A sparse transformer encoder takes as input a sparse path (i.e., reducing path length from $N$ to $N^{\prime}$), resulting in $\mathcal{O}\left(L\cdot M \cdot N^{\prime 2}\right)$ complexity; (c) \emph{LightPath} further compresses the traditional transformer in terms of layers and heads, yielding complexity $\mathcal{O}\left(L^{\prime} \cdot N^{\prime 2}\right)$, making it more scalable and lightweight than a traditional transformer encoder.}
        \label{fig:tt}
      
\end{figure*}

\paragraph{\textbf{Poor scalability w.r.t. path length}}
Since a path is a sequence of road-network edges, path representation learning benefits from models that are good at capturing sequential relationships, such as
the Transformer~\cite{DBLP:conf/nips/VaswaniSPUJGKP17}.
However, a Transformer-based method~\cite{DBLP:conf/cikm/ChenLCBLLCE21} employs a self-attention mechanism, where one edge attends to all other edges in a path in each attention, resulting in quadratic complexity, $\mathcal{O}\left( N^2\right)$, in the path length $N$.
This results in poor scalability to long paths with many edges.
%
Figure~\ref{fig:scale} gives an example of the scalability w.r.t. path length $N$, covering both memory consumption and computational cost, in terms of GPU memory~(gMem.) and Giga floating point operations per second~(GFLOPs). 
%
%
We observe that when the path length $N$ increases from 50 to 200 edges, the Transformer-based method performs poorly. 
A method that scales better w.r.t. $N$ is desirable. 

\paragraph{\textbf{Very large model size. }}
Many existing PRL models have large numbers of parameters, which restricts their use in resource-limited environments.
For example, in a Transformer-based method~\cite{DBLP:conf/cikm/ChenLCBLLCE21}, where the Transformer stacks $L$ transformer layers, each layer employs multi-head (i.e., $M$ heads) attentions. Thus, the Transformer functions like a large cuboid, with a complexity of $\mathcal{O}\left(L \cdot M \cdot N^2\right)$, as shown in Figure~\ref{fig:subfig:tte}. 
For example, Table~\ref{tab:mp} shows the numbers of parameters of Transformer-based path encoders when varying the number of layers among $12\text{, }24\text{, }48\text{, and }96$ while fixing the number heads at 8 per layer and the feature dimension of the encoder at $512$. We observe that the model parameters grow dramatically when the number of encoder layers increase, preventing the models from being deployed in resource-limited environments.
%

Moreover, models with large amounts of parameters also suffer from high storage and computational costs, which is not eco-friendly.
%
%
%
More specifically, as shown in Figure~\ref{fig:subfig:mac}, for path length $N=200$, the Transformer-based model consumes almost 3.4GiB GPU memory. 
\textbf{Proposed Solution. }To tackle the above limitations, we propose \emph{LightPath}, a lightweight and scalable path representation learning approach. To address the first limitation, we first propose a sparse auto-encoder targeting good scalability, w.r.t., path length. In particular, the introduction of sparseness reduces the path length from $N$ to $N^{\prime}$ by removing edges and returning a sparse path of length $N^{\prime}$. The sparse path is fed into a Transformer-based encoder, which reduces the complexity from $\mathcal{O}\left( L\cdot M \cdot N^{2}\right)$ to $\mathcal{O}\left( L\cdot M \cdot N^{\prime 2}\right)$. As shown in Figure~\ref{fig:subfig:ste}, this reduces a huge cuboid to a slimmer cuboid. 
To avoid information loss due to the removed edges, we connect the encoder with a decoder, with the aim of reconstructing the full path. This enables scalable yet effective unsupervised training. 
To further improve the training of the sparse encoder, we add an additional training scheme based on relational reasoning. In particular, for each path $p_i$, we construct two distinct sparse path views, denoted as $p_i ^{1}$ and $p_i ^{2}$, using different reduction ratios, e.g., removing 40\% and 80\%, respectively. 
Then, we propose a dual sparse path encoder, including the original main encoder, and an additional, auxiliary encoder. The dual sparse path encoder encodes the two path views. Thus, we achieve four path presentations $PR_i ^{1}$, $PR_i ^{2}$, $\hat{PR}_i ^{1}$, and $\hat{PR}_i ^{2}$ for path $p_i$ according to the two path views and the two sparse path encoders, where $PR$ and $\hat{PR}$ denote the representations from the main and the auxiliary encoders, respectively. Finally, given two path representations, we train a relational reasoning network to determine whether the two path representations are from the same “relation.” If they are from the same path, we consider them as positive relations; otherwise, they are negative relations.

To address the second limitation, we propose a global-local knowledge distillation framework that aims to reduce the model size of the main path encoder, which not only enables use in resource-limited environments but also improves accuracy. 
To this end, we consider the main path encoder as a teacher, and we create a lightweigth sparse encoder with fewer layers and one head as a student, further reducing a slimmer cuboid to a slim rectangle (cf. Figure (\ref{fig:subfig:lp})).
The global knowledge distillation tries to push the lightweight student to mimic the  teacher from a global semantic level (i.e., path representation level), while the local knowledge distillation can push the lightweight student to mimic the edge correlations from the teacher,
thus building a lightweight encoder while maintaining or even improving the accuracy of downstream tasks.

To the best of our knowledge, this is the first study that systematically targets lightweight and scalable path representation learning.
The study makes four main contributions.

\begin{itemize}
    \item \textit{Sparse Auto-encoder.} We propose a unified sparse auto-encoder framework that provides \emph{LightPath} with good scalability. w.r.t.\ path length.
    \item \textit{Relational Reasoning.} We introduce relational reasoning to enable efficient sparse auto-encoder training. Specifically, we propose two types of relational reasoning objectives for accurate and efficient path representation learning. These two objectives regularize each other and yield a more effective path encoder.
    \item \textit{Global-local Knowledge Distillation.} We propose a novel global-local knowledge distillation framework that enables a lightweight student sparse encoder to mimic a larger teacher sparse encoder from global and local perspectives. 
    \item \textit{Extensive Experiments.} We report on extensive experiments on two large-scale, real-world datasets with two downstream tasks. The results offer evidence of the efficiency and scalability of the proposed framework as compared with nine baselines.
\end{itemize}

\section{Preliminaries}
\label{sec:pre}
We first cover important concepts that underlie the paper's proposal and then state the problem addressed.
\subsection{Definitions}
\begin{myDef}
    \textbf{Road Network.}~A road network is defined as a graph $\mathbf{G}=(V,E)$, where $V$ is a set of vertices $v_i$ that represents road intersections and $E \subseteq V \times V$ represents a set of edges $e_i=(v_j,v_k)$ that denotes road segments.
\end{myDef}
\begin{myDef}
    \textbf{Path.}~A path $p=\langle e_1, e_2,e_3,\cdots,e_N \rangle$ is a sequence of connected edges, where $e_i \in E$ denotes an edge in path and two adjacent edges share a vertex. Next, $p$. $N$ denotes the length of path, i.e., the number of edges in $p$. We let $p.\Phi=[ 1,2,3,\cdots,N ]$ denote a sequence of orders of the edges in $p$.  
\end{myDef}
\begin{myDef}
    \textbf{Sparse Path.}~A sparse path $p^{\prime}=\langle e_i \rangle_{i \in p^{\prime}.\Omega}$ contains a subset of the edges in path $p$, where $p^{\prime}.\Omega$ is a sub-sequence of $p.\Phi$.
\end{myDef}
\noindent
\textbf{Example.} Given a path $p=\langle e_1,e_3,e_4,e_6,e_7 \rangle$ and $p.\Phi=[1,2,3,4,5]$ then path $p^{\prime}=\langle e_1,e_4,e_7 \rangle$, where $p^{\prime}.\Omega =[1,3,5]$, is one of the sparse paths for $p$.
%
%

\begin{myDef}
    \textbf{Edge Representation.} The edge representation of an edge in a road network graph is a vector in $\mathbb{R}^{d_k}$, where $d_k$ is the dimensionality of the vector. For simplicity, we reuse $e_i$ to denote an edge representation.
\end{myDef}

\begin{myDef}
    \textbf{Path Representation.} The path representation $\mathit{PR}$ of a path $p$ is a vector in $\mathbb{R}^{d}$.
\end{myDef}

\subsection{Problem Definition}
Given a set of paths $\mathbb{P}=\left\{p_{i}\right\}_{i=1}^{|\mathbb{P}|}$ in a road network $G$, scalable and efficient path representation learning aims to learn a function $SPE_{\theta}\left(\cdot \right)$ that can generate a generic path representation for each path $p_i \in \mathbb{P}$ without relying on labels. It can be formulated as follows. 

\begin{equation}
\mathit{PR} =SPE_{\theta}\left(p_{i}\right):\mathbb{R}^{N \times d_k} \rightarrow \mathbb{R}^{d}\text{ , }
\end{equation}
\noindent
where $\mathit{PR}$ is learned path representation, $\theta$ represents the learnable parameters for the sparse path encoder, $N$ is the path length and $d_k$ and $d$ are the feature dimensions for an edge and a final path representation, respectively. 

\subsection{Downstream Tasks}
A downstream task is a task that consumes a path representation. We consider travel time estimation and path ranking score estimation. In particular, we formulate task estimation as a regression problem and define the corresponding regression model as:
\begin{equation}
Reg_{task_k(\psi)}\left(\mathit{PR}_{i}\right):\mathbb{R}^{d} \rightarrow \mathbb{R}\text{ , }
\end{equation}
%
\noindent
where $task_k(\psi)$ is a learnable parameter for task $k$ and $\mathit{PR}_i$ is the learned path representation of $p_i$.

\section{Sparse Path Encoder}
\label{sec:spe}

\subsection{Transformer based Encoder}
\label{sec:sub:transformer}
We first introduce the Transformer based encoder due to its parallelism pipeline and effectiveness for long sequence modeling. Thus, given a sequence of edge representations $\mathbf{X}_p=\langle e_1, e_2, e_3, \cdots, e_N \rangle$ for a path $p$. Transformer based encoder takes as input $\mathbf{X}_p$ and returns the encoded edge representations $\mathbf{Z}_p=\langle z_1, z_2, z_3, \cdots, z_N \rangle$ that capture the correlation of different edges. Especially, instead of employing a single attention function, we define multi-head attention that linearly projects the queries, keys and values into $M$ subspaces with different, learned linear projections to $d_k$, $d_k$ and $d_v$ dimensions, respectively. Multi-head attention allows the model to jointly attend to information from different representation subspaces at different positions. Then, we formulate it as:

\begin{equation}
\mathbf{Z}_p=\mathbf{MultiHead}(\mathbf{X}_p)=\mathit{\text{Concat}}\left(\text { head }_{1}, \ldots, \text { head }_{M}\right) \cdot \mathbf{W}^{O}\text{ , } 
\end{equation}
\begin{equation}
\mathit{\text { head }_i}(\cdot)=
\operatorname{softmax}\left(\left(\mathbf{X}_p \mathbf{W}_{i}^{Q}\right) \left(\mathbf{\mathbf{X}_p W}_{i}^{K}\right)^{T}/{\sqrt{d_{k}}}\right) \left(\mathbf{\mathbf{X}_p} \mathbf{W}_{i}^{V}\right)\text{ , }
\end{equation}

where $\mathit{\text{Concat}(\cdot,\cdot)}$ represents concatenation. $\mathbf{W}_{i}^{Q} \in \mathbb{R}^{d_{model}\times d_k}$,  $\mathbf{W}_{i}^{K} \in \mathbb{R}^{d_{model}\times d_k}$, $\mathbf{W}_{i}^{V} \in \mathbb{R}^{d_{model}\times d_v}$, $\mathbf{W}^{O} \in \mathbb{R}^{Md_{v}\times d_{model} }$ are projections parameter matrices for scaled dot-product attention with respect to the learnable parameter $\theta$ in \emph{LightPaht}. $M$ denotes number of heads. $d_{model}$ represents the feature dimension of final output. $\mathbf{Z}_p 
\in \mathbb{R}^{N \times d_k}$.

Except the attention sub-layers, each of the layers in Transformer based encoder also contains a fully connected feed-forward network (FFN), which is used to each position separately and identically. This FFN consists of two linear transformations with ReLU activation in between. Specifically, we have
\begin{equation}
\operatorname{FFN}(\mathbf{Z}_p)=\max \left(0, \mathbf{Z}_p \mathbf{W_{1}^{FFN}}+\mathbf{b_{1}^{FFN}}\right) \mathbf{W_{2}^{FFN}}+\mathbf{b_{2}^{FFN}}\text{ , }
\end{equation} 
\noindent
where $\mathbf{W_{1}^{FFN}}$, $\mathbf{W_{2}^{FFN}}$, $\mathbf{b_{1}^{FFN}}$, and $\mathbf{b_{2}^{FFN}}$ are learnable parameters of feed-forward network, and $\operatorname{FFN}(\mathbf{Z}_p) 
\in \mathbb{R}^{N \times d}$.

However, Transformer based encoder suffers from poor scalability w.r.t. path length and large mode size (ref. as to Section \ref{sec:intro}). To this end, we aim to study a sparse path encoder.

\subsection{Overview}
\begin{figure}[t]
    \centering
    \includegraphics[scale=0.6]{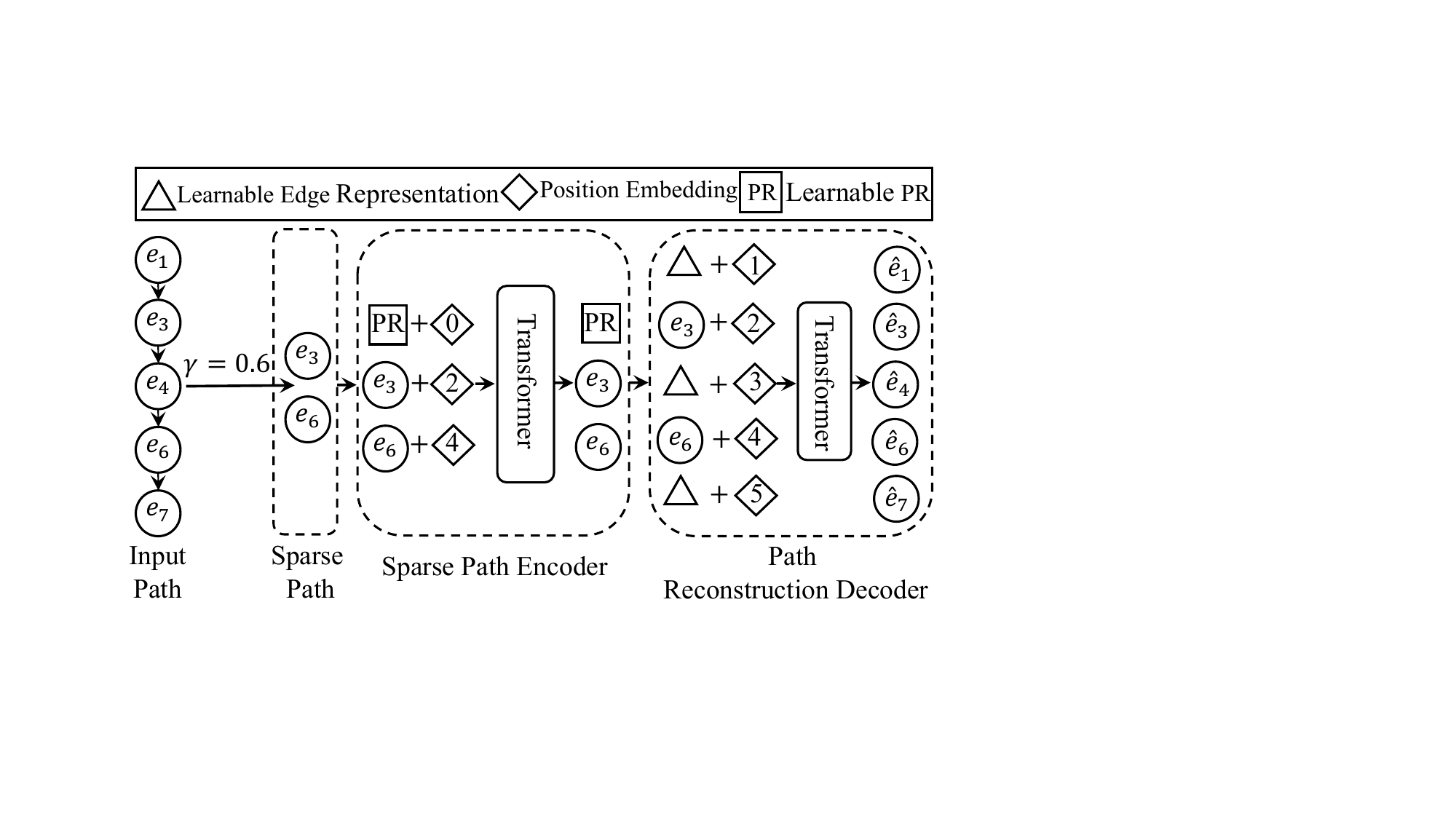}
    \caption{Sparse Auto-encoder. We remove a subset of edges from a path based on a reduction ratio $\gamma$ to obtain a sparse path. We introduce a learnable path representation in front of the sparse path. And then, we fed the resulting sparse path vectors with position embeddings to a Transformer based encoder. We then introduce a learnable edge representation, denoted as a triangle, to represent the removed edges. The encoded edges in the sparse path and the removed edge representations with position embeddings are processed by a decoder that reconstructs the edges in the original path.}
    \label{fig:spe}

\end{figure}

Figure~\ref{fig:spe} illustrates the sparse path encoder framework, which includes a sparsity operation, a sparse path encoder, and a path reconstruction decoder. The sparsity operation takes as input a full path and returns a sparse path with respect to a reduction ratio $\gamma$. Sparse path encoder takes as input a sparse path and learnable path representation and outputs learned path representations.
Next, we introduce a path reconstruction decoder to reconstruct the path, thus ensuring the learned path representation captures the entire path information.

\subsection{Sparsity Operation}
\label{sec:pr}
A path consists of a sequence of edges $p={\langle e_1, e_2,e_3,\cdots,e_N} \rangle$, which are the basic processing units of different sequential models.
The processing times of sequential models become longer when the path gets longer.
Thus,
we propose a sparsity operation, which is an approach to reduce the path length from $N$ to $N^{\prime}$, where $N^{\prime}$ is much less than $N$.
For simplicity, we conduct path reduction by randomly removing a subset of edges in a path based on a high reduction ratio $\gamma$ (e.g., $\gamma=0.6$).
A high reduction ratio $\gamma$ (the ratio for edge removal) can significantly reduce the length of each input path, thus enabling the scalability of the path.  
Specifically, we construct the sparsity operation as:
\begin{equation}
    p^{\prime}= f\left(p,\gamma \right)=\langle e_j \rangle_{j \in \Omega}\text{ , }
\end{equation}
\noindent
where $p$ is input path. $p^{\prime}$ denotes the sparse path.
For example, as shown in Figure~\ref{fig:spe}, if we have a path $p = \langle e_1,e_3,e_4,e_6,e_7 \rangle$, then we conduct sparsity operation, which randomly removes a subset of edges in $p$, i.e., $\langle e_1,e_4,e_7 \rangle$ based on reduction ratio $\gamma=0.6$ and achieve the sparse path $p^{\prime}=\langle e_3, e_6 \rangle$ and $p^{\prime}.\Omega=[2,4]$. Thus, we can reduce path from $N$ to $N^{\prime}$, i.e., from $5$ to $2$ in this example.

\subsection{Learnable Path Representation}
We use Transformer as our path encoder since it processes the input edges parallelly with respect to the self-attention mechanism. In contrast, the recurrent neural network (RNN) family is inefficient due to its recurrent nature.
To avoid achieving path representation through extra aggregation function~\cite{DBLP:conf/ijcai/YangGHT021}, we add a super extra learnable path representation 
representative $\mathit{PR}$ in front of each sparse path. Moreover, $\mathit{PR}$ is attached to position $0$ for every path, thus enabling it to capture global information of the path during the training procedure.  Thus, we update the $p^{\prime}$ as:
\begin{equation}
    p^{\prime} = \langle \mathit{PR}\rangle + \langle e_j \rangle_{j \in \Omega} =\langle e_k \rangle_{k \in \Omega^{\prime}}\text{ , }
\end{equation}
\noindent
where $e_0 = \mathit{PR}$ denotes a virtual edge and $\Omega^{\prime} = [0, \Omega]$.

To preserve the sequential information of the path, we add learnable position embedding into the sparse path representations based on order information in $\Omega^{\prime}$. Specifically, we have:
\begin{equation}
\mathbf{X}_p^{\prime}=\mathit{\text{Concat}}\langle x_k \rangle_{k \in \Omega^{\prime}}, \text{ where } x_k = e_k + pos_k  \text{ , } 
\end{equation}
\noindent
where $pos_k$ represents the learnable position embedding for edges in the sparse path and $\mathbf{X}_p^{\prime}$ represents the sparse path edge representation after concatenation. 

Take Figure~\ref{fig:spe} as an example, we first construct $p^{\prime}=\langle \mathit{PR}, e_3, e_6 \rangle$. Then, we add corresponding position embedding to the edge vectors of $p^{\prime}$, i.e., positions $0$, $2$, and $4$,
where the added position embeddings can help the Transformer encoder to be aware of the input order instead of treating them as a set of unordered path edges. Meanwhile, they enable the learned path representation $\mathit{PR}$ to capture global-level semantics in the sense that edges might play a different role in a road network. The intuition is that the super learnable path representation representative
$\mathit{PR}$ can attend attention with other edges, which captures global-level semantic. In contrast, the learnable edge representation aims to construct a full path set and reconstruct the specific edge in input path.

\subsection{Transformer Path Encoder} 
To achieve better performance, we usually stack multiple Transformer layers, each consisting of two sub-layers: multi-head attention and position-wise feed-forward network mentioned above (ref. as to Section~\ref{sec:sub:transformer} ). Motivated by~\cite{DBLP:conf/cvpr/HeZRS16}, we employ a residual connection around each sub-layers, followed by layer normalization~\cite{DBLP:journals/corr/BaKH16}. The stacked transformer model can be formulated as:
\begin{equation}
Z_p^{\prime}=\operatorname{LayerNorm}(\mathbf{X}_p^{\prime}+\text { MultiHead }(\mathbf{X}_p^{\prime}))\text{ , } 
\end{equation}
\begin{equation}
\mathit{PR}=\operatorname{LayerNorm}\left(Z_p^{\prime}+\operatorname{FFN}\left(Z_p^{\prime}\right)\right)\text{ ,}
\end{equation}
\noindent
where LayerNorm represents layer normalization and $\mathit{PR}$ is learned path representation.
Remarkably, our path encoder only takes as input a small subset of edges (e.g., 60\%) of the full path edges, which means we ignore the removed edges and just consider unremoved edges during the encoder stage to enable the path scalability. Path scalability enables us to train our path encoders concerning different lengths of path effectively and reduce the corresponding computational cost and memory usage. 
%

\subsection{Path Reconstruction Decoder}
To capture the global information of the full path, we further introduce a lightweight path decoder to reconstruct the removed edges in a path.
As shown in Figure~\ref{fig:spe}, we first complement the encoded path edges and path representation with a shared, learnable vector that represents the presence of a removed edge based on the original index of each edge in a path. Then, we add the position embedding vectors to all edge representation, which enables the learnable path representation vector to capture the global information of the entire path.
Next, the path decoder takes as input the full set of representations, including (1) path representation, (2) encoded unremoved edges, and (3) removed edges.
We select a more lightweight decoder structure, which has less number of Transformer layers.
%
%
Since the path decoder is only used to perform path reconstruction, the architecture of our path decoder can be more flexible and independent of the path encoder. Thus, the decoder is much shallower than the encoder, e.g., one layer for the decoder and 12 layers for the encoder, which significantly reduces training time. 
We reconstruct the input path by predicting the removed edges to ensure the learned path representation contains complete information about the entire path. We employ mean squared error (MSE) as our reconstruction loss function and compute MSE between the reconstructed and initial edge representations in the edge level. We only employ MSE loss on removed edges, which can be formulated as follows:
\begin{equation}
    \mathcal{L}_{rec}= \frac{1}{N}\sum_{i=1}^{N}\left(e_{i}-\hat{e_{i}} \right)^{2}\text{ , }
\end{equation}
%
\noindent
where $e_i$ and $\hat {e_{i}}$ are the initial and predicted masked edge representation, respectively. $N$ represents the number of edges for each input path.

\begin{figure}[t]
     \centering
     \begin{subfigure}[b]{0.5\textwidth}
         \centering
         \includegraphics[scale=0.45]{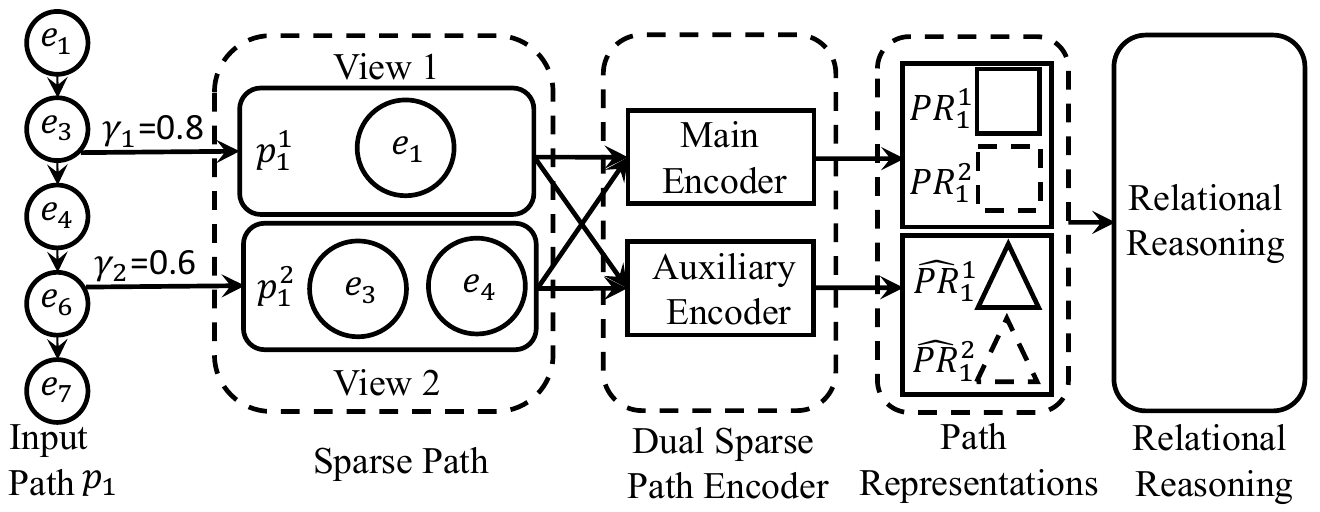}
         \caption{Path Representation Construction Using Dual Sparse Path Encoders}
         \label{fig:subfig:prc}
     \end{subfigure}
     \hfill
     \begin{subfigure}[b]{0.5\textwidth}
         \centering
         \includegraphics[scale=0.45]{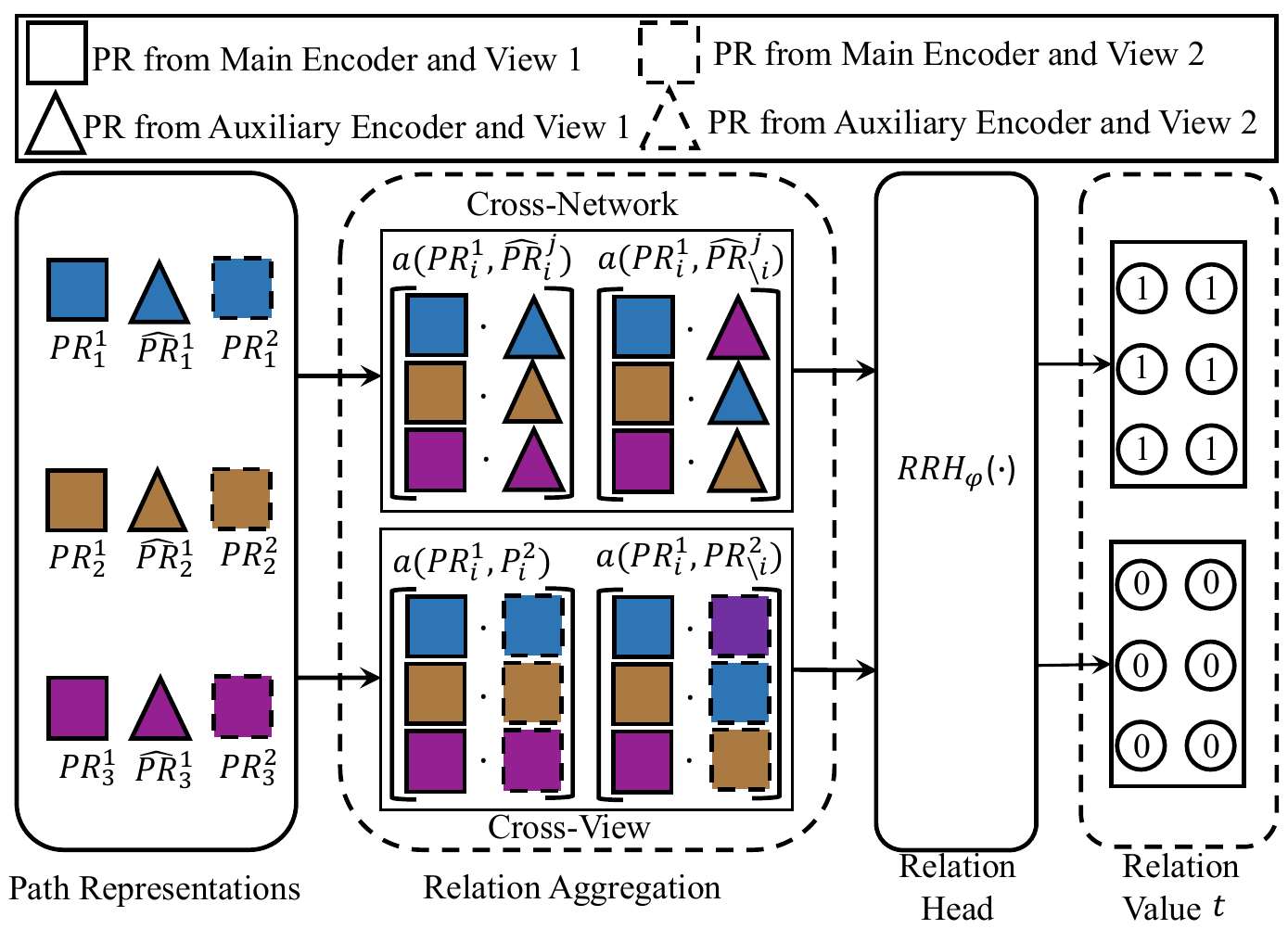}
         \caption{Relational Reasoning}
         \label{fig:subfig:msrr}
\end{subfigure}
\caption{Illustration of RR Training: (a) Given an input path $p_1$, we construct two path views (i.e., $p_1^1$ and $p_1^2$) through two reduction ratios $\gamma_1$ and $\gamma_2$, based on which a main encoder and an auxiliary encoder are employed to generate path representations for each view (i.e., $\mathit{PR}_1^1$, $\mathit{PR}_1^2$, $\hat{\mathit{PR}}_1^1$, and $\hat{\mathit{PR}}_1^2$). (b) After getting corresponding path representations for paths in a minibatch, a relational reasoning path representation learning scheme, which utilizes both cross-network and cross-view relational reasoning modules, is deployed. In particular, for both modules, an aggregation function $a$ joins positives (representations of the same paths, e.g., $a(\mathit{PR}_1^1,\hat{\mathit{PR}}_1^1)$, $a(\mathit{PR}_1^1, \mathit{PR}_1^2)$) and negatives (randomly paired representations, e.g., $a(\mathit{PR}_1^1,\hat{\mathit{PR}}_3^1)$, $a(\mathit{PR}_1^1, \mathit{PR}_3^2)$) through a commutative operator. Then relation head module $RRH_{\varphi}(\cdot)$ estimates the relation score $y$, which must be 1 for positive and 0 for negatives. Both cross-network and cross-view objectives are optimized minimizing the binary cross-entropy (BCE) between prediction and target relation value $t$. In this example, $i \in [1,2,3]$ denotes the number of paths in the minibatch and $j  \in [1,2]$ represents the number of views.}
\label{fig:msrr}

\end{figure}
%
\section{Relational Reasoning Path Representation Learning}
\label{sec:msrr}

\subsection{Overview}
To further enhance sparse auto-encoder (cf. Section~\ref{sec:spe}) training, we propose a novel self-supervised relational reasoning (RR) framework, as shown in Figure~\ref{fig:msrr}. 
The intuition behind this is that we train a relation head $RRH_{\varphi}(\cdot)$ to discriminate how path representations relate to
themselves (same class) and other paths (different class).  
In particular, this framework consists of path representation construction (cf. Figure~\ref{fig:subfig:prc}) and relational reasoning (cf. Figure~\ref{fig:subfig:msrr}), which includes cross-network relational reasoning and cross-view relational reasoning. To train our dual sparse auto-encoder, we first generate two path views, denotes as $p_1^1$ and $p_1^2$, based on two different reduction ratios $\gamma_1$ and $\gamma_2$. After this, by processing these two path views via the main encoder and the auxiliary encoder of the sparse path encoder, we construct different paths on multiple views in the representation space.
%
Finally, we employ relational reasoning to enable efficient path representation learning.

\subsection{Dual Sparse Path Encoder}
In this section, we introduce our dual sparse path encoder (SPE) that is employed to generate different path representations based on different path views. As shown in Figure \ref{fig:subfig:prc}, given a path $p_1$, we first generate sparse paths in terms of two different reduction ratios $\gamma_1$ and $\gamma_2$. We consider them as different path views, i.e., path view 1 and path view 2. Then, our dual sparse path encoder, including a main encoder and an auxiliary encoder, takes as input two different path views (i.e., $p_1^{1}$ and $p_1^{2}$) and returns different path representations. Specifically, each encoder takes as input two different path views and returns two different path representations, where solid and dotted $\square$ denote the path representations returned from main encoder based on path view 1 and path view 2, respectively, i.e., $\mathit{PR}_1^{1}$ and $\mathit{PR}_1^{2}$. In contrast, solid and dotted $\triangle$ represent the path representations achieved from auxiliary encoder based on both path views, respectively, i.e,. $\hat{\mathit{PR}}_1^{1}$ and $\hat{\mathit{PR}}_1^{2}$. To this end, we construct four different path representations for a given path, which promote our design of cross-network relational reasoning and cross-view relational reasoning in turn. Finally, we formulate it as:
\begin{equation}
    \mathit{PR}_i^{j} =SPE_{\theta}(p_i^{j},\gamma)\text{ , } \hat{\mathit{PR}}_i^{j} =SPE_{\hat{\theta}}(p_i^{j},\gamma)\text{ , }
\end{equation}
\noindent
where $\mathit{PR}_i^{j}$ and $\hat{\mathit{PR}}_i^{j}$ are path representations obtained from the main encoder and the auxiliary encoder, respectively. $p_i$ denotes the $i$-th path in the path set. $j \in [1,2]$ denotes the path views. $\theta$ and $\hat{\theta}$ are the parameters for the main encoder and auxiliary encoder.

\subsection{Relational Reasoning}

\subsubsection{Cross-Network Relational Reasoning}
\label{subsec:cn}
In \emph{LightPath}, we employ a dual sparse path encoder, which includes main and auxilary encoder, as shown in Figure~\ref{fig:subfig:prc}. We first construct path representations through sparsity operation based on different reduction ratios $\gamma_1$ and $\gamma_2$. Given a set of path $\bigl \{p_1,p_2,\cdots, p_K\bigr \}$, we can have a set of path representations $\bigl \{ \mathit{PR}_{1}^{1}, \mathit{PR}_{2}^{1}, \cdots, \mathit{PR}_{K}^{1} \bigr \}$ from main encoder and $\bigl \{ \hat{\mathit{PR}}_{1}^{1},\hat{\mathit{PR}}_{2}^{1},\cdots, \hat{\mathit{PR}}_{K}^{1} \bigr \}$ or $\bigl \{ \hat{\mathit{PR}}_{1}^{2},\hat{\mathit{PR}}_{2}^{2},\cdots,\hat{\mathit{PR}}_{K}^{2} \bigr \}$ from auxiliary encoder by using path representation construction. Then we employ relation aggregation $a(\cdot)$ that joins the positive path representation relations $\langle \mathit{PR}_{i}^{1}$, $\hat{\mathit{PR}}_{i}^{1} \rangle$ or $\langle \mathit{PR}_{i}^{1}$, $\hat{\mathit{PR}}_{i}^{2} \rangle$ and the negative path representation relations $\langle \mathit{PR}_{i}^{1}$, $\hat{\mathit{PR}}_{\backslash i}^{1} \rangle$, where $i$ denotes the $i$-th path sample and $\backslash i \neq i$ represents randomly selected path representations in a minibatch. Take Figure~\ref{fig:subfig:msrr} as an example, where $K=3$. we join $\langle \mathit{PR}_{1}^{1}$, $\hat{\mathit{PR}}_{1}^{1} \rangle$ as a positive relation pair (representation from same path), and $\langle \mathit{PR}_{1}^{1}$, $\hat{\mathit{PR}}_{2}^{1} \rangle$ as a negative relation pair (representation from different paths) through aggregation function $a$.
Next, the relational head $RRH_{\varphi}(\cdot)$, which is non-linear function approximator parameterized by $\varphi$, takes as input representation relation pairs of cross-network and returns a relation score $y$. Finally, we formulate the cross-network relational reasoning task as a binary classification task, where we use binary cross-entropy loss to train our sparse path encoder, which is given as follows.


\begin{equation}
\label{eq:loss_cn}
\mathcal{L}_{cn}=\underset{\boldsymbol{\theta}, \varphi}{\operatorname{argmin}} \sum_{i=1}^{K}\sum_{j=1}^{2}  \mathcal{L}\left(RRH_{\varphi}\left(a\left(\mathit{PR}_{i}^{j}, \hat{\mathit{PR}}_{i}^{j}\right)\right), t=1\right) +  \mathcal{L}\left(RRH_{\varphi}\left(a\left(\mathit{PR}_{i}^{j}, \hat{\mathit{PR}}_{\backslash i}^{j}\right)\right), t=0\right)\text{ , }
\end{equation}

\noindent
where $K$ is the the number of path samples in the minibatch. $a(\cdot,\cdot)$ is an aggregation function. $\mathcal{L}$ is a loss function between relation score and a target relation value. $t$ is a target relation values.

The intuition behind this is to discriminate path presentations of same path and different paths, which are from different views across dual sparse path encoder and are able to distill the knowledge from historical observations, as well as stabilizing the main encoder training. 
To realize this, we adopt Siamese architecture for our dual sparse path encoder, where the auxiliary encoder does not directly receive the gradient during the training procedure. In contrast, we update its parameters by leveraging the momentum updating principle:


\begin{equation}
\label{eq:momentum}
    \hat{\theta}^{t}= m \times \hat{\theta}^{(t-1)} +(1-m)\times \theta^{t},
\end{equation}
\noindent
where $m$ is momentum parameters. $\theta$ and $\hat{\theta}$ are the parameters of the main encoder and the auxiliary encoder.

\subsubsection{Cross-View Relational Reasoning}

To enhance the learning ability of our \emph{LightPath}, we further consider the ties between two views within main encoder, which acts as a strong requarization to enhance the learning ability of our methods. We do not have to include such relational reasoning within the auxiliary encoder because it will not directly compute gradient during training, and our goal is to train main encoder.
Figure~\ref{fig:subfig:msrr} shows the design of our cross-view relational reasoning, which contains two similar representations from two views based on $\gamma_1$ and $\gamma_2$. 
The intuition of cross-view relational reasoning is to preserve the relation between two views of the same path and discriminate them from the view of other paths. 

Similar with cross-network, given a set of paths $\bigl \{ p_1, p_2,\cdots, p_K \bigr \}$.
We first achieve two set of path representations in terms of two path views based on main encoder, i.e., $\bigl \{ \mathit{PR}_1^{1}, \mathit{PR}_2^{1},\cdots,\mathit{PR}_k^{1} \bigr \}$ and $\bigl \{ \mathit{PR}_1^{2}, \mathit{PR}_2^{2},\cdots,PR_K^{2} \bigr \}$. Then, we join the positive relation pairs (e.g., $\langle \mathit{PR}_i^{1}, \mathit{PR}_i^{2} \rangle$) and negative relation pairs (e.g., $\langle \mathit{PR}_i^{1}, \mathit{PR}_{\backslash{i}}^{2} \rangle$) through aggregation function. For example, as shown in Figure~\ref{fig:subfig:msrr}, there are 3 paths in the set. Thus, we can denote $\langle \mathit{PR}_1^{1}, \mathit{PR}_1^{2} \rangle$ as a positive pair and $\langle \mathit{PR}_1^{1}, \mathit{PR}_3^{2} \rangle$ as a negative pair.
Then, we further employ relational head $RRH_{\varphi}(\cdot)$, which takes as input a positive pair and a negative pairs from different views, to achieve the corresponding relation score $y$ for the cross-view relational reasoning.
Last, we formulate the cross-view relational reasoning loss to discriminate how different views of a  path is related to themselves and other paths. In this phase, the complete learning objective can be specified as:

\begin{equation}
\label{eq:loss_cv}
\mathcal{L}_{cv}=\underset{\mathbf{\theta},\mathbf{\varphi}}{\operatorname{argmin}} \sum_{i=1}^{K}  \mathcal{L}\left(RRH_{\varphi}\left(a\left(\mathit{PR}_{i}^{1}, \mathit{PR}_{i}^{2}\right)\right), t=1\right) +\mathcal{L}\left(RRH_{\varphi}\left(a\left(\mathit{PR}_{i}^{1}, \mathit{PR}_{\backslash i}^{2}\right)\right), t=0\right)\text{ , }
\end{equation}
\noindent
where $K$ is the the number of path samples in the minibatch.

\subsubsection{Objective for \emph{RR}}

To train our dual path encoder end-to-end and efficient learn path representations for downstream tasks, we jointly leverage both the cross-network and cross-view relation reasoning loss. Specifically, the overall objective function is formulated as Eq.~\ref{eq:msrr}.

\begin{equation}
\label{eq:msrr}
    \min_{\theta, \varphi}\mathcal{L}_{RR}=\mathcal{L}_{cn} +\mathcal{L}_{cv}
\end{equation}

\subsection{LightPath Training}
To train our sparse path encoder and learn path representations for downstream tasks, we jointly minimize the reconstruction and RR loss. Specifically, the overall objective function is defined as:

\begin{equation}
\label{eq:loss_}
    \mathcal{L}=\mathcal{L}_{rec}+\mathcal{L}_{RR}
\end{equation}

%

\section{Global Local Knowledge Distillation (GLKD)}

\begin{figure}
    \centering
    \includegraphics[scale=0.6]{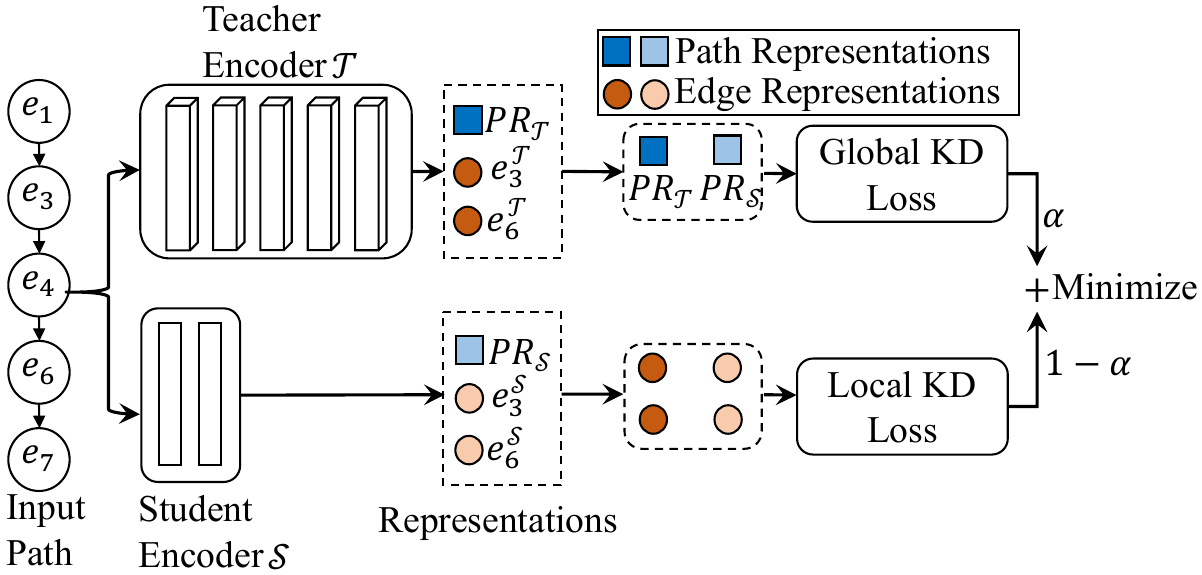}
    \caption{Illustration of GLKD. Given an input path, we formulate our GLKD as a weighted sum of global path representation knowledge distillation (GPRKD) loss and local edge representation knowledge distillation (LERKD) loss.}
    \label{fig:kd}

\end{figure}

So far, we realize our \emph{LightPath} through sparse auto-encoder and relational reasoning and transform it from large cuboid (cf. Figure~\ref{fig:subfig:tte} to a slim cuboid (cf. Figure~\ref{fig:subfig:ste}). To enable the \emph{LightPath} that can be deployed on resource-limited mobile devices, we introduce our global-local knowledge distillation (GLKD) to further reduce the size of the sparse auto-encoder, as shown in Figure~\ref{fig:kd}. We first train a large cuboid teacher encoder with multiple transformer layers and heads (cf. Figure~\ref{fig:subfig:ste}) based on path reconstruction (cf. Section~\ref{sec:spe}) and relational reasoning (cf. Section~\ref{sec:msrr} ). Then, we employ a small rectangle student encoder (cf. Figure~\ref{fig:subfig:lp}), which has less layers and heads, to mimic a large teacher model and use the teacher's knowledge to obtain similar or superior performance based on GLKD. 
Specifically, GLKD constructs a local knowledge distillation by matching the representations of correlated edges. On such a basis, the global term distills the knowledge from teacher to student that enabling the informative and powerful path representation for the student model. 

\subsection{Global-path Representation Distillation}
%

%

Given a path $p_i=\langle e_1,e_2,e_3,\cdots,e_N\rangle$, where $N$ is the number of edges in a path. We define $\mathit{PR}^\mathcal{T}(p_{i})$ and $\mathit{PR}^{\mathcal{S}}(p_{i})$ represent the path representations achieved from teacher encoder $\mathcal{T}_{\theta}$ and student encoder $\mathcal{S}_{\theta}$. The intuition of global path representation knowledge distillation is to let the student encoder mimic the global properties captured by a large cuboid teacher encoder. And thus, the goal of global path representation knowledge distillation is to put closer the path representation from teacher encoder and student encoder in the latent space. We formalize this problem as minimizing a latent space distance  representation pairs in terms of the large cuboid teacher encoder and the rectangle student encoder. The formulation of the objective function is given as follows: 
\begin{equation}
\small
    \min_{\theta}\mathcal{L}_{global}\left(\mathbf{\mathit{PR}}^{\mathcal{T}}({p_{i}}), \mathbf{\mathit{PR}}^{\mathcal{S}}(p_{i})\right)=\left\|sp(\mathbf{\mathit{PR}}^{\mathcal{T}}(p_{i})/t)-sp(\mathbf{\mathit{PR}}^{\mathcal{S}}(p_{i})/t)\right\|^{2}\textbf{,}
\end{equation}
\noindent
where $sp(\cdot)$ is exponential function. $t$ denotes the temperature. Using a higher value for $t$ produces a softer probability distributions over path representations.

\subsection{Local-edge Correlation Distillation}
The goal of local-edge structure distillation is to preserve the local similarity of the edge correlations in a path. In particular, it is expected that the representation of the same edge in a path represented by the teacher encoder and the student encoder should be close to each other. The intuition is that a rectangle student encoder mimics the edge correlations in a path captured by a large cuboid teacher encoder. Using a similarity measurement, we formulate the local-edge structure distillation problem as minimizing the latent space distance of edge representations from the teacher encoder and then student encoder. 

In specific, given a path $p = \langle e_1,e_2,e_3,\cdots e_N \rangle$ in a road network, where $N$ is the number of edges in a path. Through applying an $L$-layers Transformer encoder (i.e., teacher encoder $\mathcal{T}_{\theta}$) and $L^{\prime}$-layers Transformer encoder (i.e., student encoder $\mathcal{S}_{\theta}$) upon sparse path $p^{\prime}$, where $L \ll L^{\prime}$, the edge representation that captures spatial dependencies are derived as follows. 
\begin{equation}
    F^{\mathcal{T}}(e_{i})^{N^{\prime}}_{i=1} = \mathcal{T}_{\theta}(p)\text{ , } F^{\mathcal{S}}(e_{i})^{N^{\prime}}_{i=1} = \mathcal{S}_{\theta}(p)\text{ , }
\end{equation}
\noindent
where $F^{\mathcal{T}}(e_{i})^{N^{\prime}}_{i=1}$ and $F^{\mathcal{S}}(e_{i})^{N^{\prime}}_{i=1}$ represent the edge representation with respect to the teacher and student encoder, respectively.

In this phase, the goal of learning is to reduce the latent space distance between same edge pair from the teacher and student encoder, respectively. To this end, we aim to minimize the following objective functions between edge representation pairs in terms of the parameters of the student encoder.

\begin{equation}
\small
    \min_{\theta}\mathcal{L}_{local}\left(\mathbf{F}^{\mathcal{T}}(e_{i}), \mathbf{F}^{\mathcal{S}}(e_{i})\right)=\frac{1}{n}\sum^{n}_{i=1}\left\|sp(\mathbf{F}^{\mathcal{T}}(e_{i})/t)-sp(\mathbf{F}^{\mathcal{S}}(e_{i})/t)\right\|^{2}\textbf{,}
\end{equation}
\noindent
where $sp(\cdot)$ represents exponential function. $t$ denotes the temperature. Using a higher value for $t$ produces a softer probability distributions over edges.

\subsection{Objective for \emph{GLKD}}

To train our global and local knowledge distillation in an end-to-end fashion, we jointly leverage both the global and local knowledge distillation loss. Specifically, the overall objective function to minimize is defined in Eq.~\ref{eq:ad}.

\begin{equation}
\label{eq:ad}
\min_{\theta}\mathcal{L}_{GLKD}=\alpha*\mathcal{L}_{global} +\left(1-\alpha\right)*\mathcal{L}_{local}\text{ , }
\end{equation}
\noindent
where $\alpha$ is balancing factor.

\section{Experiments}
\subsection{Experimental Setup}
\subsubsection{Datasets} We conduct experiments on two real-world datasets and one synthetic dataset to enable fair comparisons with existing studies. Based on two real-world datasets, we report results for two downstream tasks: travel time estimation~\cite{DBLP:journals/corr/abs-2203-16110,DBLP:conf/sigmod/Yuan0BF20}, and path ranking~\cite{DBLP:journals/pvldb/0026HFZL020, DBLP:conf/icde/Yang020,yang2020context}. Due to the lack of large amounts of long paths in the real-world datasets, we construct one synthetic dataset that contains paths with lengths of 100, 150, and 200 to verify the efficiency and scalability of \emph{LightPath}.
%

\noindent
\textbf{Aalborg, Denmark:} We collect the road network of Aalborg from OpenStreetMap\footnote{\href{https://www.openstreetmap.org}{https://www.openstreetmap.org}} that contains 10,017 nodes and 11,597 edges. Specifically, this dataset contains 180 million GPS records from 183 vehicles sampled at 1 Hz over a two-year period from 2017 to 2018. After map matching~\cite{DBLP:conf/gis/NewsonK09}, we obtain 39,160 paths with length 50. 
%

\noindent
\textbf{Chengdu, China:} This dataset was collected from Chengdu, China, on October and November 2016. We obtain the corresponding road network from OpenStreetMap. The network contains 6,632 nodes and 17,038 edges. The GPS data was sampled at about 1/4-1/2 Hz. We obtain 50,000 paths through map matching with lengths 50.

%
\noindent
\textbf{Synthetic: }Due to the lack of large amounts of long paths in the real-world datasets, we construct one synthetic dataset to verify the efficiency and scalability of \emph{LightPath}. In particular, we first randomly pick 500 nodes in road network of Aalborg dataset, and then expand each node to a path by random walking until the path length reach the threshold (i.e., 100, 150, 200), which we refer as to generation process. Subsequently, we iterate the generation process 10 times to construct 5,000 paths for each path length.    

\subsubsection{Downstream Tasks} We report the results on two downstream tasks: 

\noindent
\textbf{Path Travel Time Estimation: }We obtain travel time (in seconds) for each path from the trajectory. We aim to utilize a regression model to predict the travel time based on the learned path representations. We employ Mean Absolute Error(\textbf{MAE}), Mean Absolute Relative Error(\textbf{MARE}), and Mean Absolute Percentage Error(\textbf{MAPE}) to evaluate the performance of travel time estimations. The smaller values of these metrics, the better performance we achieve.

\noindent
\textbf{Path Ranking: }Each path is assigned a ranking score in the range $[0,1]$, which is obtained from historical trajectories by following the existing studies~\cite{yang2020context,DBLP:conf/icde/Yang020}. 
More specifically, we take the path that is used by a driver in the historical trajectories as the trajectory path, which is denoted as the top ranked path.
Then, we 
generate multiple paths connecting the same source and destination via path finding algorithms~\cite{DBLP:journals/tkde/LiuJYZ18}. Finally, we calculate the similarity between a generated path and the trajectory path as a ranking score. The higher ranking score indicates a generated path is more similar to the trajectory path, and the trajectory path itself has a score of 1 and ranks the highest. 
To measure the path ranking, we apply \textbf{MAE}, the Kendall rank correlation coefficient ($\tau$), and the Spearman's rank correlation coefficient ($\rho$), which are widely used metrics in path ranking, to evaluate the effectiveness of path ranking.

\subsubsection{Models for Downstream Tasks}
For all unsupervised learning methods, we first achieve the corresponding $d$ dimensionality path representation and then we build a regression model that takes as input a path representation and output estimated the travel time and path ranking, respectively.
In particular, we select ensemble model \textit{Gradient Boosting Regressor}(GBR)~\cite{DBLP:conf/nips/PeterDHN17} as our prediction model since they are regression problems.

\subsubsection{Baselines}
We compare~\emph{LightPath} with 9 baselines, including 6 unsupervised learning-based methods and 3 supervised learning-based methods. 
The details of these baseline methods are summarized as follows:
\begin{itemize}
    \item \textbf{Node2vec}~\cite{DBLP:conf/kdd/GroverL16} is an unsupervised node representation model that learn node representation in a graph. We achieve the path representation by aggregating the node representations of the nodes in a path. 
    
    \item \textbf{MoCo}~\cite{DBLP:conf/cvpr/He0WXG20} is a momentum contrast for unsupervised visual representation learning. Here we use momentum contrast to learn path representations.
    \item \textbf{Toast}~\cite{DBLP:conf/cikm/ChenLCBLLCE21} first uses auxiliary traffic context information to learn road segment representation based on the skip-gram model and then utilizes a stacked transformer encoder layer to train trajectory representation through route recovery and trajectory discrimination tasks. We use the same schema to learn path representations.
    
    \item \textbf{t2vec}~\cite{DBLP:conf/icde/LiZCJW18} is a trajectory representation learning method for similarity computation based on the encoder-decoder framework, which is trained to reconstruct the original trajectory. We use a sequence of edges in a path to represent a trajectory. 
    \item \textbf{NeuTraj}~\cite{DBLP:conf/icde/YaoCZB19} is a method that revised the structure of LSTM to
    learn representations of the grid in the process of training their
    framework. To support our task with it, we replace the grid with edges in their framework.
    \item \textbf{PIM}~\cite{DBLP:conf/ijcai/YangGHT021} is an unsupervised path representation learning approach that first generates negative samples using curriculum learning and then employs global and local mutual information maximization to learn path representations.
    \item \textbf{HMTRL}~\cite{DBLP:journals/pvldb/0026HFZL020} is a supervised path representation learning framework for multi-modal transportation recommendation. 
    \item \textbf{PathRank}~\cite{yang2020context} is a supervised path representation learning model based on GRUs, which treats departure time and driver ID as additional information.
    
    \item \textbf{CompactETA}~\cite{DBLP:conf/kdd/FuMYW20} aims to estimate travel time based on a real time predictor and an asynchronous updater.  \item \textbf{HierETA}~\cite{DBLP:conf/kdd/ChenXGFMCC22} is a supervised multi-view trajectory representation method to estimate the travel time. 
    \item \textbf{LightPath-Sup} is a supervised version of our \emph{LightPath}, where we train it in a supervised manner.
\end{itemize}

	For all unsupervised learning methods, we first use unlabeled training data (e.g., 30K unlabeled Aalborg dataset and 50K unlabeled Chengdu dataset) to train path encoders to obtain path representations. Then a regression model takes as input path representations and returns the estimated travel time and path ranking score using a limited labeled training dataset, e.g., the 12K labeled Aalborg dataset. In comparison, for all supervised learning methods, we directly train path encoders using a limited labeled training dataset, e.g., the 12K labeled Aalborg dataset and Chengdu dataset.
	
\subsubsection{Implementation Details}
We employ an asymmetrical sparse auto-encoder architecture and randomly initialize all learnable parameters with uniform distributions. 
In particular, we adopt Siamese architecture, where we update the parameters of the auxiliary encoder based on the momentum updating principle based on the main encoder and we set the momentum parameter $m=0.99$.
We employ node2vec~\cite{DBLP:conf/kdd/GroverL16} to embed each edge to 128-dimensional vectors and set the dimension for path representation to 128. For a fair comparison, we set the path representation dimensionality of all baseline methods as 128.
%
%
We select concatenate as the relation aggregation function $a(\cdot,\cdot)$.
We use the AdamW optimizer with a cosine decay learning rate schedule over 400 epochs, with a warm-up period of 40 epochs. We set the base learning rate to 1e-3 and betas as $(0.9,0.95)$.
We vary $\gamma$ from 0.1,0.3,0.5,0.7,0.9 to study the effect of path scalability and efficiency for the \emph{LightPath}.
In addition, we consider four different path lengths, i.e., 50, 100, 150, and 200, to study the effectiveness, efficiency, and scalability of the \emph{LightPath}. 
We then evaluate our \emph{LightPath} as well as all baselines on a powerful Linux server with 40 Intel(R) Xeon(R) W-2155 CPU @ 3.30GHz and two TITAN RTX GPU cards. Finally, all algorithms are implemented in PyTorch 1.11.0.

\subsection{Experimental Results}
\subsubsection{Overall Performance}
Table~\ref{tab:ttpr} shows the overall performance of our \emph{LightPath} and all the compared baselines on both datasets in terms of different evaluation metrics. Especially, we select 30K unlabeled paths on Aalborg and Chengdu, respectively, but we only have 12K labeled paths for both datasets. 
\begin{table*}[!htp]
	\caption{{Overall Accuracy on Travel Time Estimation and Ranking Score Estimation}}
	\centering
	\small
	
	\begin{tabular}{l|llllll|llllll}
		\toprule[2pt]
		\multirow{3}{*}{\textbf{Method}} & \multicolumn{6}{l|}{\textbf{Aalborg}}                                                                                                                & \multicolumn{6}{l}{\textbf{Chengdu}}                                                                                                                \\ \cline{2-13} 
		& \multicolumn{3}{l|}{\textbf{Travel Time Estimation}}                                 & \multicolumn{3}{l|}{\textbf{Path Ranking}}                   & \multicolumn{3}{l|}{\textbf{Travel Time Estimation}}                                 & \multicolumn{3}{l}{\textbf{Path Ranking}}                   \\ \cline{2-13} 
		& \multicolumn{1}{l|}{\textbf{MAE}}    & \multicolumn{1}{l|}{\textbf{MARE}} & \multicolumn{1}{l|}{\textbf{MAPE}}  & \multicolumn{1}{l|}{\textbf{MAE}}  & \multicolumn{1}{l|}{\textbf{$\tau$}}  & \textbf{$\rho$}  & \multicolumn{1}{l|}{\textbf{MAE}}    & \multicolumn{1}{l|}{\textbf{MARE}} & \multicolumn{1}{l|}{\textbf{MAPE}}  & \multicolumn{1}{l|}{\textbf{MAE}}  & \multicolumn{1}{l|}{\textbf{$\tau$}}  & \textbf{$\rho$}  \\ \toprule[1pt]
		\emph{Node2vec}                & \multicolumn{1}{l|}{154.07} & \multicolumn{1}{l|}{0.20} & \multicolumn{1}{l|}{25.22} & \multicolumn{1}{l|}{0.24} & \multicolumn{1}{l|}{0.59} & 0.64 & \multicolumn{1}{l|}{267.28} & \multicolumn{1}{l|}{0.23} & \multicolumn{1}{l|}{26.30} & \multicolumn{1}{l|}{0.15} & \multicolumn{1}{l|}{0.74} & 0.77 \\ \hline
		\emph{MoCo}                    & \multicolumn{1}{l|}{146.29} & \multicolumn{1}{l|}{0.19} & \multicolumn{1}{l|}{21.60} & \multicolumn{1}{l|}{0.25} & \multicolumn{1}{l|}{0.53} & 0.57 & \multicolumn{1}{l|}{237.14} & \multicolumn{1}{l|}{0.20} & \multicolumn{1}{l|}{23.13} & \multicolumn{1}{l|}{0.15} & \multicolumn{1}{l|}{0.77} & 0.81 \\ \hline
		\emph{Toast}                   & \multicolumn{1}{l|}{137.27} & \multicolumn{1}{l|}{0.17} & \multicolumn{1}{l|}{20.43} & \multicolumn{1}{l|}{0.24} & \multicolumn{1}{l|}{0.59} & 0.63 & \multicolumn{1}{l|}{240.57} & \multicolumn{1}{l|}{0.21} & \multicolumn{1}{l|}{23.50} & \multicolumn{1}{l|}{0.11} & \multicolumn{1}{l|}{0.65} & 0.68 \\ \hline
		\emph{t2vec}                            & \multicolumn{1}{l|}{147.24} & \multicolumn{1}{l|}{0.19} & \multicolumn{1}{l|}{22.13} & \multicolumn{1}{l|}{0.25} & \multicolumn{1}{l|}{0.52} & 0.56 & \multicolumn{1}{l|}{242.96} & \multicolumn{1}{l|}{0.21} & \multicolumn{1}{l|}{23.65} & \multicolumn{1}{l|}{0.14} & \multicolumn{1}{l|}{0.77} & 0.82 \\ \hline
		\emph{NeuTraj}                          & \multicolumn{1}{l|}{117.06} & \multicolumn{1}{l|}{0.15} & \multicolumn{1}{l|}{18.09} & \multicolumn{1}{l|}{0.25} & \multicolumn{1}{l|}{0.60} & 0.64 & \multicolumn{1}{l|}{232.96} & \multicolumn{1}{l|}{0.20} & \multicolumn{1}{l|}{22.73} & \multicolumn{1}{l|}{0.12} & \multicolumn{1}{l|}{0.79} & 0.83 \\ \hline
		\emph{PIM}                     & \multicolumn{1}{l|}{102.09} & \multicolumn{1}{l|}{0.14} & \multicolumn{1}{l|}{14.92} & \multicolumn{1}{l|}{0.20} & \multicolumn{1}{l|}{0.63} & 0.67 & \multicolumn{1}{l|}{223.34} & \multicolumn{1}{l|}{0.19} & \multicolumn{1}{l|}{21.69} & \multicolumn{1}{l|}{0.12} & \multicolumn{1}{l|}{0.80} & 0.84 \\ \hline
		\emph{HMTRL}                   & \multicolumn{1}{l|}{101.81} & \multicolumn{1}{l|}{0.13} & \multicolumn{1}{l|}{14.51} & \multicolumn{1}{l|}{0.17} & \multicolumn{1}{l|}{0.68} & 0.72 & \multicolumn{1}{l|}{218.94} & \multicolumn{1}{l|}{0.19} & \multicolumn{1}{l|}{21.22} & \multicolumn{1}{l|}{0.09} & \multicolumn{1}{l|}{0.83} & 0.84 \\ \hline
		\emph{PathRank}                         & \multicolumn{1}{l|}{115.37} & \multicolumn{1}{l|}{0.15} & \multicolumn{1}{l|}{16.41} & \multicolumn{1}{l|}{0.21} & \multicolumn{1}{l|}{0.64} & 0.68 & \multicolumn{1}{l|}{229.85} & \multicolumn{1}{l|}{0.20} & \multicolumn{1}{l|}{22.53} & \multicolumn{1}{l|}{0.11} & \multicolumn{1}{l|}{0.81} & 0.82 \\ \hline
		\emph{CompactETA}                         & \multicolumn{1}{l|}{106.47} & \multicolumn{1}{l|}{0.15} & \multicolumn{1}{l|}{16.22} & \multicolumn{1}{l|}{0.17} & \multicolumn{1}{l|}{0.67} & 0.70 & \multicolumn{1}{l|}{236.28} & \multicolumn{1}{l|}{0.20} & \multicolumn{1}{l|}{23.13} & \multicolumn{1}{l|}{0.11} & \multicolumn{1}{l|}{0.79} & 0.80 \\ \hline
		\emph{HierETA}                         & \multicolumn{1}{l|}{\textit{88.95}} & \multicolumn{1}{l|}{\textit{0.12}} & \multicolumn{1}{l|}{\textit{14.23}} & \multicolumn{1}{l|}{\textit{0.15}} & \multicolumn{1}{l|}{\textit{0.71}} & \textit{0.74} & \multicolumn{1}{l|}{\textit{215.39}} & \multicolumn{1}{l|}{\textit{0.19}} & \multicolumn{1}{l|}{\textit{21.66}} & \multicolumn{1}{l|}{\textit{0.09}} & \multicolumn{1}{l|}{\textit{0.84}} & \textit{0.85} \\ \hline
		\emph{LightPath-Sup}           & \multicolumn{1}{l|}{105.51} & \multicolumn{1}{l|}{0.15} & \multicolumn{1}{l|}{16.35} & \multicolumn{1}{l|}{0.14} & \multicolumn{1}{l|}{0.68} & 0.72 & \multicolumn{1}{l|}{218.67} & \multicolumn{1}{l|}{0.19} & \multicolumn{1}{l|}{21.36} & \multicolumn{1}{l|}{0.13} & \multicolumn{1}{l|}{0.76} & 0.79 \\ \hline
		\emph{LightPath}                        & \multicolumn{1}{l|}{\textbf{85.76}}  & \multicolumn{1}{l|}{\textbf{0.11}} & \multicolumn{1}{l|}{\textbf{12.12}} & \multicolumn{1}{l|}{\textbf{0.13}} & \multicolumn{1}{l|}{\textbf{0.73}} & \textbf{0.77} & \multicolumn{1}{l|}{\textbf{212.61}} & \multicolumn{1}{l|}{\textbf{0.18}} & \multicolumn{1}{l|}{\textbf{20.75}} & \multicolumn{1}{l|}{\textbf{0.07}} & \multicolumn{1}{l|}{\textbf{0.87}} & \textbf{0.88} \\ \toprule[2pt]
	\end{tabular}
	\label{tab:ttpr}
	
\end{table*}
Thus, we use 30K unlabeled paths to train path encoder for unsupervised-based methods. However, supervised approaches can only use the 12K labeled paths.  
Overall, \emph{LightPath} outperforms all the baselines on these two tasks for both datasets, which demonstrates the advance of our model. 
Specifically, we can make the following observations. Graph representation learning based approach \emph{Node2vec} is much worse than \emph{LightPath}. This is because \emph{Node2vec} fails to capture spatial dependencies in a path. In contrast, \emph{LightPath} considers the spatial dependencies through the self-attention mechanism, thus achieving better performance.

Although \emph{MoCo} considers the dependencies among edges in a path, this method still performs worse. The main reason is that \emph{MoCo} can leverage the spatial dependencies, but it converges very slow since it needs large amounts of negative samples to enable training.
\emph{LightPath} also outperform \emph{t2vec} and \emph{NeuTraj}, which both are first design to learn trajectory representation for trajectory similarity computation. This suggests that random drops on some edges and not reconstruct these edges in a path resulting in spatial information missing, thus achieving the worse performance on downstream tasks. 
\emph{PIM} consistently outperforms all other unsupervised baselines, which demonstrates the effectiveness of representation learning. The main reason is that \emph{PIM} is designed for path representation learning. However, \emph{PIM} is InfoNCE based method and has high computation complexity,
making it hard to deploy on resource-limited edge devices.
%
%
\emph{HMTRL}, \emph{PathRank} and \emph{LightPath-Sup} are three supervised learning methods that achieve relatively worse performance due to the lack of labeled training data. Since labeling data
is very time-consuming and expensive. We consider a scenario where labeled data is limited in this paper.

\begin{figure}[t]
	\centering
	\begin{subfigure}[b]{0.45\textwidth}
		\centering
		\includegraphics[width=\textwidth]{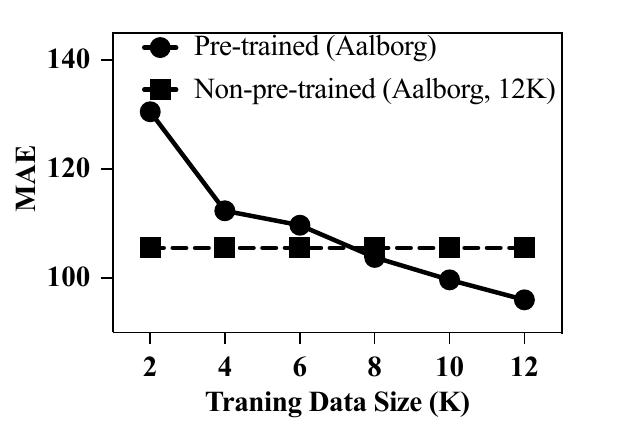} 
		\caption{Travel Time Estimation}
		\label{fig:subfig:tteaal}
	\end{subfigure}
	\hfill
	\begin{subfigure}[b]{0.45\textwidth}
		\centering
		\includegraphics[width=\textwidth]{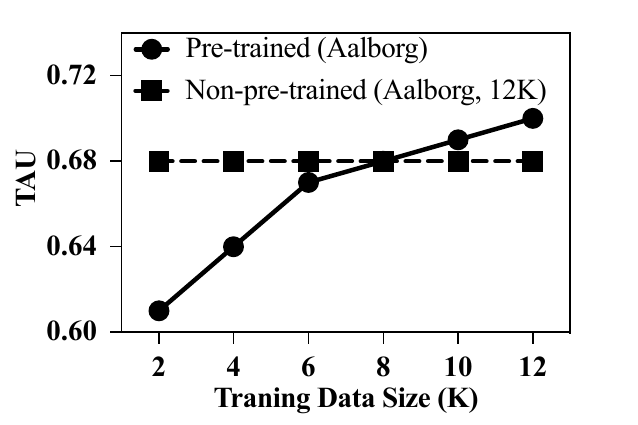}
		\caption{Path Ranking Estimation}
		\label{fig:subfig:pr}
	\end{subfigure}
	
	\begin{subfigure}[b]{0.45\textwidth}
		\centering
		\includegraphics[width=\textwidth]{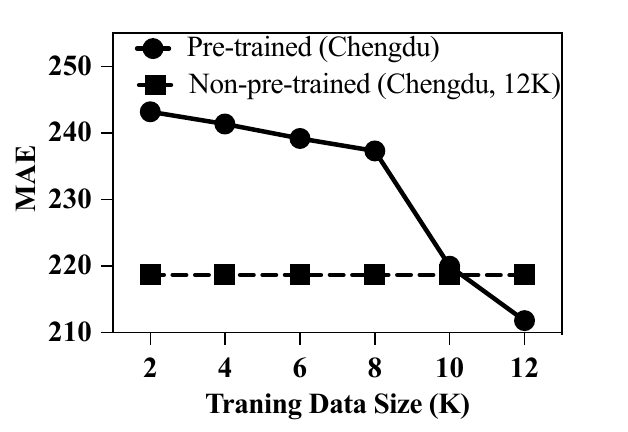}
		\caption{Travel Time Estimation}
		\label{fig:subfig:ttecd}
	\end{subfigure}
	\hfill
	\begin{subfigure}[b]{0.45\textwidth}
		\centering
		\includegraphics[width=\textwidth]{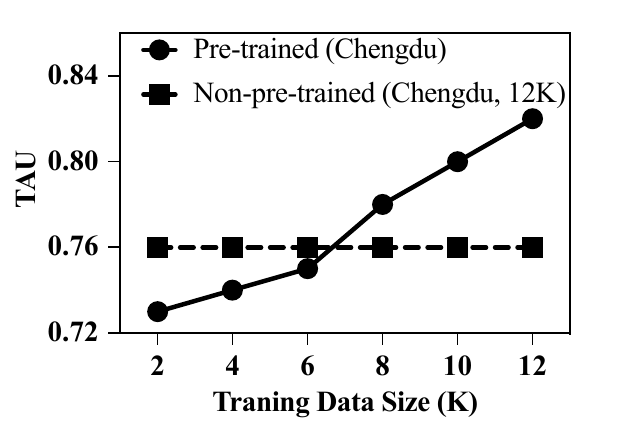}
		\caption{Path Ranking Estimation}
		\label{fig:subfig:prcd}
	\end{subfigure}
	
	\caption{Effects of Pre-training.}
	\label{fig:pretrain}
	
\end{figure}
\subsubsection{Using \emph{LightPath} as Pre-training Methods}
Model pre-training aims to create generic representations that can then be fine-tuned for multiple downstream tasks using a limited labeled dataset. Especially for many personalized services (e.g., personalized travel time estimation or path ranking estimation) in transportation applications, a user can use his/her trajectory to fine-tune the pre-trained model and then achieve the personalized service, where we do not have many trajectories from the user.
%
In this experiment, we evaluate the effect of Pre-training. We employ \emph{LightPath} as a pre-training method for the supervised method \emph{LightPath-Sup}. Specifically, we first train \emph{LightPath} in an unsupervised fashion, and then we use the learned transformer path encoder to initialize the transformer in \emph{LightPath-Sup}. Here, it takes as input a sequence of edge representations and predicts the travel time and path ranking score. 
Figure~\ref{fig:pretrain} illustrates the performance of \emph{LightPath-Sup} w. and w/o pre-training over two downstream tasks on both datasets. When employing non-pre-trained \emph{LightPath-Sup}, we train it using 12K labeled training paths. We notice that (1) when employing pre-training, we can obtain the same performance with no-pre-trained \emph{LightPath-Sup} using less labeled data. For example, 
\emph{LightPath-Sup} w. pre-training only needs 8K, and 10K labeled training paths for the Aalborg and Chengdu, respectively, to achieve the same performance of \emph{LightPath-Sup} w/o pre-training with 12k labeled samples on the task of travel time estimation. (2) 
\emph{LightPath-Sup} w. pre-training achieves higher performance than
\emph{LightPath-Sup} w/o pre-training. We observe similar results on the task of path ranking, demonstrating that \emph{LightPath} can be used as a pre-training method to enhance supervised methods.

%
\begin{table}[t]
\centering
\small
\caption{Effect of Variants of \emph{LightPath}}
\begin{tabular}{l|llllll|llllll}
\toprule[2pt]
\multirow{3}{*}{\textbf{Method}} & \multicolumn{6}{l|}{\textbf{Aalborg}}                                                                                                                                                              & \multicolumn{6}{l}{\textbf{Chengdu}}                                                                                                                                                              \\ \cline{2-13} 
                                 & \multicolumn{3}{l|}{\textbf{Travel Time Estimation}}                                                        & \multicolumn{3}{l|}{\textbf{Path Ranking}}                                           & \multicolumn{3}{l|}{\textbf{Travel Time Estimation}}                                                        & \multicolumn{3}{l}{\textbf{Path Ranking}}                                           \\ \cline{2-13} 
                                 & \multicolumn{1}{l|}{\textbf{MAE}} & \multicolumn{1}{l|}{\textbf{MARE}} & \multicolumn{1}{l|}{\textbf{MAPE}} & \multicolumn{1}{l|}{\textbf{MAE}} & \multicolumn{1}{l|}{\textbf{$\tau$}} & \textbf{$\rho$} & \multicolumn{1}{l|}{\textbf{MAE}} & \multicolumn{1}{l|}{\textbf{MARE}} & \multicolumn{1}{l|}{\textbf{MAPE}} & \multicolumn{1}{l|}{\textbf{MAE}} & \multicolumn{1}{l|}{\textbf{$\tau$}} & \textbf{$\rho$} \\ \toprule[1pt]
\textit{\textbf{w/o RR}}         & \multicolumn{1}{l|}{90.90}        & \multicolumn{1}{l|}{0.12}          & \multicolumn{1}{l|}{13.85}         & \multicolumn{1}{l|}{0.17}         & \multicolumn{1}{l|}{0.66}         & 0.70         & \multicolumn{1}{l|}{224.31}       & \multicolumn{1}{l|}{0.19}          & \multicolumn{1}{l|}{21.90}         & \multicolumn{1}{l|}{0.14}         & \multicolumn{1}{l|}{0.76}         & 0.79         \\ \hline
\textit{\textbf{w/o Rec.}}       & \multicolumn{1}{l|}{103.45}       & \multicolumn{1}{l|}{0.14}          & \multicolumn{1}{l|}{15.76}         & \multicolumn{1}{l|}{0.15}         & \multicolumn{1}{l|}{0.65}         & 0.69         & \multicolumn{1}{l|}{229.24}       & \multicolumn{1}{l|}{0.20}          & \multicolumn{1}{l|}{22.36}         & \multicolumn{1}{l|}{0.16}         & \multicolumn{1}{l|}{0.69}         & 0.73         \\ \hline
\textit{\textbf{w/o ME}}         & \multicolumn{1}{l|}{91.57}        & \multicolumn{1}{l|}{0.12}          & \multicolumn{1}{l|}{13.09}         & \multicolumn{1}{l|}{0.16}         & \multicolumn{1}{l|}{0.68}         & 0.72         & \multicolumn{1}{l|}{223.81}       & \multicolumn{1}{l|}{0.19}          & \multicolumn{1}{l|}{21.86}         & \multicolumn{1}{l|}{0.13}         & \multicolumn{1}{l|}{0.78}         & 0.81         \\ \hline
\textit{\textbf{w/o CN}}         & \multicolumn{1}{l|}{93.17}        & \multicolumn{1}{l|}{0.12}          & \multicolumn{1}{l|}{13.35}         & \multicolumn{1}{l|}{0.15}         & \multicolumn{1}{l|}{0.68}         & 0.73         & \multicolumn{1}{l|}{217.14}       & \multicolumn{1}{l|}{0.18}          & \multicolumn{1}{l|}{21.29}         & \multicolumn{1}{l|}{0.08}         & \multicolumn{1}{l|}{0.80}         & 0.83         \\ \hline
\textit{\textbf{w/o CV}}         & \multicolumn{1}{l|}{89.84}        & \multicolumn{1}{l|}{0.12}          & \multicolumn{1}{l|}{13.51}         & \multicolumn{1}{l|}{0.15}         & \multicolumn{1}{l|}{0.68}         & 0.72         & \multicolumn{1}{l|}{215.59}       & \multicolumn{1}{l|}{0.18}          & \multicolumn{1}{l|}{21.20}         & \multicolumn{1}{l|}{0.09}         & \multicolumn{1}{l|}{0.81}         & 0.83         \\ \hline
\textit{\textbf{LightPath}}      & \multicolumn{1}{l|}{\textbf{85.76}} & \multicolumn{1}{l|}{\textbf{0.11}} & \multicolumn{1}{l|}{\textbf{12.12}} & \multicolumn{1}{l|}{\textbf{0.13}} & \multicolumn{1}{l|}{\textbf{0.73}} & \textbf{0.77} & \multicolumn{1}{l|}{\textbf{212.61}} & \multicolumn{1}{l|}{\textbf{0.18}} & \multicolumn{1}{l|}{\textbf{20.75}} & \multicolumn{1}{l|}{\textbf{0.07}} & \multicolumn{1}{l|}{\textbf{0.87}} & \textbf{0.88} \\ \toprule[2pt]
\end{tabular}
\label{tab:basicl}
\end{table}

\subsubsection{Ablation Studies}
To verify the effectiveness of different components in \emph{LightPath}, we conduct ablation studies on \emph{LightPath}: a) effect of variants of \emph{LightPath}, specifically reconstruction (Rec) loss, relational reasoning (RR) loss, cross-network loss and cross-view loss; b) effect of global-local knowledge distillation.

\textbf{a)} \textit{Effect of variants of \emph{LightPath}}, we consider five variants of \emph{LightPath}: 1) \textit{w/o RR}; 2) \textit{w/o Rec.}; 3) \textit{w/o ME}; 
4) \textit{w/o CN}; 5) \textit{w/o CV}. 
In \textit{w/o RR}, we only consider path reconstruction loss and use main encoder; In \textit{w/o Rec.}, we only consider relational reasoning loss;  In \textit{w/o ME}, we consider both path reconstruction and relational reasoning losses, but we do not consider Siamese architectures in dual path encoder; 
In \textit{w/o CN}, we remove the cross-network loss in RR; And in \textit{w/o CV}, we remove cross-view loss in RR.
%
Table~\ref{tab:basicl} reports the results on both dataset. We can observe that (1) \emph{LightPath} \textit{w/o Rec.} achieves the worst performance because the learned $\mathit{PR}$ only capture information from sparse path while ignoring the removed edges, which verifies the importance of path reconstruction decoder; (2) \emph{LightPath} \textit{w/o RR} also achieves the poor performance, which implies the effectiveness of self-supervised relational reasoning.
(3) We observe that the performance of \emph{LightPath} degrades without cross-network and cross-view loss on both datasets, which further demonstrates the effectiveness of our relational reasoning loss.
(4) We notice that \emph{LightPath} achieves the best performance. This result implies
that all the proposed modules contribute positively to the final
performance, which validates that \emph{LightPath} takes advantage of all designed components.
%

\begin{table}[t]
\centering
\small
\caption{Effect of KD, Global Loss and Local Loss}
\begin{tabular}{l|llllll|llllll}
\toprule[2pt]
\multirow{3}{*}{\textbf{Method}} & \multicolumn{6}{l|}{\textbf{Aalborg}}                                                                                                                                                                    & \multicolumn{6}{l}{\textbf{Chengdu}}                                                                                                                                                                     \\ \cline{2-13} 
                                 & \multicolumn{3}{l|}{\textbf{Travel Time Estimation}}                                                           & \multicolumn{3}{l|}{\textbf{Path Ranking}}                                              & \multicolumn{3}{l|}{\textbf{Travel Time Estimation}}                                                            & \multicolumn{3}{l}{\textbf{Path Ranking}}                                              \\ \cline{2-13} 
                                 & \multicolumn{1}{l|}{\textbf{MAE}}   & \multicolumn{1}{l|}{\textbf{MARE}} & \multicolumn{1}{l|}{\textbf{MAPE}}  & \multicolumn{1}{l|}{\textbf{MAE}}  & \multicolumn{1}{l|}{\textbf{$\tau$}}  & \textbf{$\rho$}  & \multicolumn{1}{l|}{\textbf{MAE}}    & \multicolumn{1}{l|}{\textbf{MARE}} & \multicolumn{1}{l|}{\textbf{MAPE}}  & \multicolumn{1}{l|}{\textbf{MAE}}  & \multicolumn{1}{l|}{\textbf{$\tau$}}  & \textbf{$\rho$}  \\ \toprule[1pt]
\textit{\textbf{w/o KD}}         & \multicolumn{1}{l|}{87.77}          & \multicolumn{1}{l|}{0.11}          & \multicolumn{1}{l|}{12.94}          & \multicolumn{1}{l|}{0.14}          & \multicolumn{1}{l|}{0.70}          & 0.74          & \multicolumn{1}{l|}{213.26}          & \multicolumn{1}{l|}{0.18}          & \multicolumn{1}{l|}{20.97}          & \multicolumn{1}{l|}{0.08}          & \multicolumn{1}{l|}{0.84}          & 0.86          \\ \hline
\textit{\textbf{w/o Global}}     & \multicolumn{1}{l|}{90.24}          & \multicolumn{1}{l|}{0.12}          & \multicolumn{1}{l|}{13.31}          & \multicolumn{1}{l|}{0.18}          & \multicolumn{1}{l|}{0.67}          & 0.71          & \multicolumn{1}{l|}{220.32}          & \multicolumn{1}{l|}{0.19}          & \multicolumn{1}{l|}{21.52}          & \multicolumn{1}{l|}{0.09}          & \multicolumn{1}{l|}{0.79}          & 0.81          \\ \hline
\textit{\textbf{w/o Local}}      & \multicolumn{1}{l|}{89.23}          & \multicolumn{1}{l|}{0.12}          & \multicolumn{1}{l|}{12.78}          & \multicolumn{1}{l|}{0.16}          & \multicolumn{1}{l|}{0.69}          & 0.73          & \multicolumn{1}{l|}{215.03}          & \multicolumn{1}{l|}{0.19}          & \multicolumn{1}{l|}{21.02}          & \multicolumn{1}{l|}{0.08}          & \multicolumn{1}{l|}{0.82}          & 0.84          \\ \hline
\textit{\textbf{LightPath}}      & \multicolumn{1}{l|}{\textbf{85.76}} & \multicolumn{1}{l|}{\textbf{0.11}} & \multicolumn{1}{l|}{\textbf{12.12}} & \multicolumn{1}{l|}{\textbf{0.13}} & \multicolumn{1}{l|}{\textbf{0.73}} & \textbf{0.77} & \multicolumn{1}{l|}{\textbf{212.61}} & \multicolumn{1}{l|}{\textbf{0.18}} & \multicolumn{1}{l|}{\textbf{20.75}} & \multicolumn{1}{l|}{\textbf{0.07}} & \multicolumn{1}{l|}{\textbf{0.87}} & \textbf{0.88} \\ \toprule[2pt]
\end{tabular}
\label{tab:glkd}
\end{table}

\begin{sidewaystable*}[htp]
\centering
\caption{Model Scalability vs. Reduction Ratio~($\gamma$) and Path Length~(N)(Aalborg) }
\small
\begin{tabular}{l|lllllllllllll}
\toprule[2pt]
\multirow{3}{*}{$N$} & \multicolumn{13}{l}{\textbf{\emph{LightPath}}}                                                                                                                           \\ \cline{2-14} 
                             & \multicolumn{2}{l|}{$\gamma=0$}                                      & \multicolumn{2}{l|}{$\gamma=0.1$}                                                                              & \multicolumn{2}{l|}{$\gamma=0.3$}                                                                              & \multicolumn{2}{l|}{$\gamma=0.5$}                                                                              & \multicolumn{2}{l|}{$\gamma=0.7$}                                                                              & \multicolumn{2}{l|}{$\gamma=0.9$}                                                                              & \multirow{2}{*}{\begin{tabular}[c]{@{}l@{}}Para.\\ \end{tabular}} \\ \cline{2-13}
                             & \multicolumn{1}{l|}{GFLOPs}  & \multicolumn{1}{l|}{\begin{tabular}[c]{@{}l@{}}gMem.\\ \end{tabular}} & \multicolumn{1}{l|}{GFLOPs} & \multicolumn{1}{l|}{\begin{tabular}[c]{@{}l@{}}gMem.\\ \end{tabular}} & \multicolumn{1}{l|}{GFLOPs} & \multicolumn{1}{l|}{\begin{tabular}[c]{@{}l@{}}gMem.\\ \end{tabular}} & \multicolumn{1}{l|}{GFLOPs} & \multicolumn{1}{l|}{\begin{tabular}[c]{@{}l@{}}gMem.\\ \end{tabular}} & \multicolumn{1}{l|}{GFLOPs} & \multicolumn{1}{l|}{\begin{tabular}[c]{@{}l@{}}gMem.\\ \end{tabular}} & \multicolumn{1}{l|}{GFLOPs} & \multicolumn{1}{l|}{\begin{tabular}[c]{@{}l@{}}gMem.\end{tabular}} &                                                                           \\ \toprule[1pt]
$ 50$                           & \multicolumn{1}{l|}{8.01}   & \multicolumn{1}{l|}{1.47}    & \multicolumn{1}{l|}{7.48}  & \multicolumn{1}{l|}{1.39}                                               & \multicolumn{1}{l|}{6.44}  & \multicolumn{1}{l|}{1.37}                                               & \multicolumn{1}{l|}{5.39}  & \multicolumn{1}{l|}{1.36}                                               & \multicolumn{1}{l|}{4.34}  & \multicolumn{1}{l|}{1.34}                                               & \multicolumn{1}{l|}{3.18}  & \multicolumn{1}{l|}{1.33}                                               & 1.570                                                                     \\ \hline
$ 100$                          & \multicolumn{1}{l|}{15.77}  & \multicolumn{1}{l|}{1.60}    & \multicolumn{1}{l|}{14.72} & \multicolumn{1}{l|}{1.50}                                               & \multicolumn{1}{l|}{12.62} & \multicolumn{1}{l|}{1.48}                                               & \multicolumn{1}{l|}{10.52} & \multicolumn{1}{l|}{1.46}                                               & \multicolumn{1}{l|}{8.43}  & \multicolumn{1}{l|}{1.44}                                               & \multicolumn{1}{l|}{6.23}  & \multicolumn{1}{l|}{1.43}                                               & 1.570                                                                     \\ \hline
$ 150$                          & \multicolumn{1}{l|}{23.53}  & \multicolumn{1}{l|}{1.72}    & \multicolumn{1}{l|}{21.95} & \multicolumn{1}{l|}{1.64}                                               & \multicolumn{1}{l|}{18.81} & \multicolumn{1}{l|}{1.60}                                               & \multicolumn{1}{l|}{15.66} & \multicolumn{1}{l|}{1.58}                                               & \multicolumn{1}{l|}{12.52} & \multicolumn{1}{l|}{1.55}                                               & \multicolumn{1}{l|}{9.27}  & \multicolumn{1}{l|}{1.52}                                               & 1.570                                                                     \\ \hline
$ 200$                          & \multicolumn{1}{l|}{31.29}  & \multicolumn{1}{l|}{1.90}    & \multicolumn{1}{l|}{29.19}  & \multicolumn{1}{l|}{1.81}                                               & \multicolumn{1}{l|}{24.99} & \multicolumn{1}{l|}{1.77}                                               & \multicolumn{1}{l|}{20.80}  & \multicolumn{1}{l|}{1.73}                                               & \multicolumn{1}{l|}{16.61} & \multicolumn{1}{l|}{1.68}                                               & \multicolumn{1}{l|}{12.31} & \multicolumn{1}{l|}{1.65}                                               & 1.570                                                                     \\ \toprule[2pt]
\multirow{3}{*}{} & \multicolumn{13}{l}{\textbf{\textit{LightPath w/o KD}}}     \\ \cline{2-14} 
                             & \multicolumn{2}{l|}{$\gamma=0$}                                      & \multicolumn{2}{l|}{$\gamma=0.1$}                                                                              & \multicolumn{2}{l|}{$\gamma=0.3$}                                                                              & \multicolumn{2}{l|}{$\gamma=0.5$}                                                                              & \multicolumn{2}{l|}{$\gamma=0.7$}                                                                              & \multicolumn{2}{l|}{$\gamma=0.9$}                                                                              & \multirow{2}{*}{\begin{tabular}[c]{@{}l@{}}Para.\\ \end{tabular}}                                           \\ \cline{2-13}
                             & \multicolumn{1}{l|}{GFLOPs}  & \multicolumn{1}{l|}{\begin{tabular}[c]{@{}l@{}}gMem.\\ \end{tabular}} & \multicolumn{1}{l|}{GFLOPs} & \multicolumn{1}{l|}{\begin{tabular}[c]{@{}l@{}}gMem.\\ \end{tabular}} & \multicolumn{1}{l|}{GFLOPs} & \multicolumn{1}{l|}{\begin{tabular}[c]{@{}l@{}}gMem.\\ \end{tabular}} & \multicolumn{1}{l|}{GFLOPs} & \multicolumn{1}{l|}{\begin{tabular}[c]{@{}l@{}}gMem.\\ \end{tabular}} & \multicolumn{1}{l|}{GFLOPs} & \multicolumn{1}{l|}{\begin{tabular}[c]{@{}l@{}}gMem.\\ \end{tabular}} & \multicolumn{1}{l|}{GFLOPs} & \multicolumn{1}{l|}{\begin{tabular}[c]{@{}l@{}}gMem.\\ \end{tabular}} &                                                                           \\ \hline
$ 50$                           & \multicolumn{1}{l|}{33.68}  & \multicolumn{1}{l|}{1.78}    & \multicolumn{1}{l|}{30.64} & \multicolumn{1}{l|}{1.70}                                               & \multicolumn{1}{l|}{22.55} & \multicolumn{1}{l|}{1.61}                                               & \multicolumn{1}{l|}{18.47} & \multicolumn{1}{l|}{1.53}                                               & \multicolumn{1}{l|}{12.39} & \multicolumn{1}{l|}{1.47}                                               & \multicolumn{1}{l|}{5.70}  & \multicolumn{1}{l|}{1.41}                                               & 5.525                                                                     \\ \hline
$ 100$                          & \multicolumn{1}{l|}{66.60}  & \multicolumn{1}{l|}{2.53}    & \multicolumn{1}{l|}{60.52}  & \multicolumn{1}{l|}{2.37}                                               & \multicolumn{1}{l|}{48.36} & \multicolumn{1}{l|}{2.11}                                               & \multicolumn{1}{l|}{36.19} & \multicolumn{1}{l|}{1.86}                                               & \multicolumn{1}{l|}{24.03}  & \multicolumn{1}{l|}{1.72}                                               & \multicolumn{1}{l|}{11.26} & \multicolumn{1}{l|}{1.58}                                               & 5.525                                                                     \\ \hline
$ 150$                          & \multicolumn{1}{l|}{99.53}  & \multicolumn{1}{l|}{3.44}    & \multicolumn{1}{l|}{90.41} & \multicolumn{1}{l|}{3.23}                                               & \multicolumn{1}{l|}{72.16}  & \multicolumn{1}{l|}{2.76}                                               & \multicolumn{1}{l|}{53.91} & \multicolumn{1}{l|}{2.35}                                               & \multicolumn{1}{l|}{35.65}  & \multicolumn{1}{l|}{2.03}                                               & \multicolumn{1}{l|}{16.82} & \multicolumn{1}{l|}{1.82}                                               & 5.525                                                                     \\ \hline
$ 200$                          & \multicolumn{1}{l|}{132.54} & \multicolumn{1}{l|}{4.74}    & \multicolumn{1}{l|}{120.29} & \multicolumn{1}{l|}{4.30}                                               & \multicolumn{1}{l|}{95.96} & \multicolumn{1}{l|}{3.53}                                               & \multicolumn{1}{l|}{71.64} & \multicolumn{1}{l|}{2.94}                                               & \multicolumn{1}{l|}{47.31} & \multicolumn{1}{l|}{2.43}                                               & \multicolumn{1}{l|}{22.37} & \multicolumn{1}{l|}{2.10}                                               & 5.525                                                                     \\ \toprule[2pt]
\end{tabular}
\label{tab:msca}
\end{sidewaystable*}

\textbf{b)} \textit{Effect of KD, global KD loss, local KD loss}: We further study the effect of global-local knowledge distillation. We compared our framework with three variants: 1) \textit{w/o KD}, which denotes the performance of the teacher model; 2) \textit{w/o global KD loss}, which removes global loss from global-local knowledge distillation; and 3) \textit{w/o local KD loss}, which removes local loss from global-local knowledge distillation. As shown in Table~\ref{tab:glkd}, compared with \textit{KD}, \emph{LightPath} achieves a better performance, which verifies that the teacher model can improve the performance of the student model. Both global and local loss can improve the performance of the learned path representation of the student model. In specific, global loss makes more contributions to the learned path representations. As a result, removing global loss degrades performance significantly.

\subsubsection{Parameter Sensitivity Analysis} We proceed to study three important hyper-parameters, including 1) model scalability w.r.t. reduction ratio and path length, 2) Effect of Reduction Ratio $\gamma$, 3) the parameter of temperature for global-local knowledge distillation, and 4) effect of balancing factor $\alpha$.

\paragraph{Model Scalability} In the sequel, we explore the model scalability in terms of reduction ratio and path length based on the synthetic dataset. Table~\ref{tab:msca} depicts the results for both \emph{LightPath} and its teacher model, with varying $\gamma = 0,0.1,0.3,0.5,0.7,0.9$. $\gamma=0$ denotes we do not conduct sparsity operation for the input path, i.e., using a classic Transformer based encoder. We can observe that the GFLOPs and gMem. (GiB) decrease with the increase in the reduction ratio. It is because the higher value of $\gamma$ is, the more edges we can remove. 
Second, \emph{LightPath} has significantly reduced model complexity, w.r.t., GFLOPs and gMem.. For example, we can reduce the training GFLOPs by $2.54 \times$ for the \emph{LightPath} by increasing the reduction ratio $\gamma$ from $0$ to $0.9$ in terms of path length 200. 
Moreover, \emph{LightPath} also shows better performance (i.e., GFLOPs and gMem.) over teacher model, e.g., $1.79 \times$ GFLOPs speedup with reduction ratio $\gamma=0.9$. 
Third, the parameters (Para. (Millions)) of teacher model is at least $3.5 \times$ of \emph{LightPath}, which implies the effectiveness of our proposed framework.
Overall, \emph{LightPath} shows potential of scalability to support path representation learning for long paths.

\begin{table}[t]
\centering
\caption{Effect of Reduction Ratio $\gamma$}
\begin{tabular}{l|llllll}
\toprule[2pt]
\multirow{3}{*}{\textbf{$\gamma$}}                                                  & \multicolumn{6}{l}{\textbf{Aalborg}}                                                                                                                                                              \\ \cline{2-7} 
                                                                                  & \multicolumn{3}{l|}{\textbf{Travel Time Estimation}}                                                        & \multicolumn{3}{l}{\textbf{Path Ranking}}                                           \\ \cline{2-7} 
                                                                                  & \multicolumn{1}{l|}{\textbf{MAE}} & \multicolumn{1}{l|}{\textbf{MARE}} & \multicolumn{1}{l|}{\textbf{MAPE}} & \multicolumn{1}{l|}{\textbf{MAE}} & \multicolumn{1}{l|}{\textbf{$\tau$}} & \textbf{$\rho$} \\ \toprule[1pt]
0.1                                                 & \multicolumn{1}{l|}{\textbf{82.79}}        & \multicolumn{1}{l|}{\textbf{0.11}}          & \multicolumn{1}{l|}{11.95}         & \multicolumn{1}{l|}{\textbf{0.12}}         & \multicolumn{1}{l|}{\textbf{0.74}}         &\textbf{0.77}          \\ \hline
0.3                                                 & \multicolumn{1}{l|}{84.75}        & \multicolumn{1}{l|}{0.11}          & \multicolumn{1}{l|}{12.14}         & \multicolumn{1}{l|}{0.13}         & \multicolumn{1}{l|}{0.73}         &0.77          \\ \hline
0.5                                                 & \multicolumn{1}{l|}{84.81}        & \multicolumn{1}{l|}{0.11}          & \multicolumn{1}{l|}{\textbf{11.86}}         & \multicolumn{1}{l|}{0.14}         & \multicolumn{1}{l|}{0.72}         &0.76          \\ \hline
0.7                                                 & \multicolumn{1}{l|}{85.91}        & \multicolumn{1}{l|}{0.11}          & \multicolumn{1}{l|}{12.49}         & \multicolumn{1}{l|}{0.14}         & \multicolumn{1}{l|}{0.71}         &0.75          \\ \hline

0.9                                &                      \multicolumn{1}{l|}{85.76}        & \multicolumn{1}{l|}{0.11}          & \multicolumn{1}{l|}{12.12}         & \multicolumn{1}{l|}{0.13}         & \multicolumn{1}{l|}{0.73}         &0.77 \\ \toprule[2pt]        
\end{tabular}
\label{tab:gamma}

\end{table}

\noindent
\paragraph{Effect of Reduction Ratio $\gamma$} 
To study the impact of reduction ratio $\gamma$ in the final performance, we conduct an experiment by varying the $\gamma$ from 0.1 to 0.9 on Aalborg, which is shown in Table~\ref{tab:gamma}. We can observe that the overall performance in both downstream tasks degrades a little when $\gamma$ increases, which is reasonable as the the model has more input information. However, we can also observe the performance differences are not so significant, which suggests the effectiveness of our proposed framework. Even when a high reduction ratio is applied, the performance does not does not go down too much. Therefore, our proposed method can achieve good scalability while ensuring accuracy. 
\begin{table}[t]
\centering
\caption{Effect of Temperature $t$ in KD}
\begin{tabular}{l|llllll|llllll}
\toprule[2pt]
\multirow{3}{*}{t} & \multicolumn{6}{l|}{\textbf{Aalborg}}                                                                                                                                                                    & \multicolumn{6}{l}{\textbf{Chengdu}}                                                                                                                                                                     \\ \cline{2-13} 
                   & \multicolumn{3}{l|}{\textbf{Travel Time Estimation}}                                                           & \multicolumn{3}{l|}{\textbf{Path Ranking}}                                              & \multicolumn{3}{l|}{\textbf{Travel Time Estimation}}                                                            & \multicolumn{3}{l}{\textbf{Path Ranking}}                                              \\ \cline{2-13} 
                   & \multicolumn{1}{l|}{\textbf{MAE}}   & \multicolumn{1}{l|}{\textbf{MARE}} & \multicolumn{1}{l|}{\textbf{MAPE}}  & \multicolumn{1}{l|}{\textbf{MAE}}  & \multicolumn{1}{l|}{\textbf{$\tau$}}  & \textbf{$\rho$}  & \multicolumn{1}{l|}{\textbf{MAE}}    & \multicolumn{1}{l|}{\textbf{MARE}} & \multicolumn{1}{l|}{\textbf{MAPE}}  & \multicolumn{1}{l|}{\textbf{MAE}}  & \multicolumn{1}{l|}{\textbf{$\tau$}}  & \textbf{$\rho$}  \\ \toprule[1pt]
1                  & \multicolumn{1}{l|}{90.01}          & \multicolumn{1}{l|}{0.12}          & \multicolumn{1}{l|}{13.30}          & \multicolumn{1}{l|}{0.16}          & \multicolumn{1}{l|}{0.65}          & 0.69          & \multicolumn{1}{l|}{225.70}          & \multicolumn{1}{l|}{0.20}          & \multicolumn{1}{l|}{22.02}          & \multicolumn{1}{l|}{0.09}          & \multicolumn{1}{l|}{0.79}          & 0.82          \\ \hline
3                  & \multicolumn{1}{l|}{94.11}          & \multicolumn{1}{l|}{0.12}          & \multicolumn{1}{l|}{13.54}          & \multicolumn{1}{l|}{0.15}          & \multicolumn{1}{l|}{0.68}          & 0.72          & \multicolumn{1}{l|}{217.07}          & \multicolumn{1}{l|}{0.19}          & \multicolumn{1}{l|}{20.77}          & \multicolumn{1}{l|}{0.09}          & \multicolumn{1}{l|}{0.81}          & 0.84          \\ \hline
5                  & \multicolumn{1}{l|}{90.39}          & \multicolumn{1}{l|}{0.12}          & \multicolumn{1}{l|}{12.76}          & \multicolumn{1}{l|}{0.15}          & \multicolumn{1}{l|}{0.66}          & 0.70          & \multicolumn{1}{l|}{216.93}          & \multicolumn{1}{l|}{0.19}          & \multicolumn{1}{l|}{21.24}          & \multicolumn{1}{l|}{0.08}          & \multicolumn{1}{l|}{0.83}          & 0.86          \\ \hline
7                  & \multicolumn{1}{l|}{89.64}          & \multicolumn{1}{l|}{0.12}          & \multicolumn{1}{l|}{12.76}          & \multicolumn{1}{l|}{0.15}          & \multicolumn{1}{l|}{0.70}          & 0.74          & \multicolumn{1}{l|}{214.88}          & \multicolumn{1}{l|}{0.19}          & \multicolumn{1}{l|}{20.98}          & \multicolumn{1}{l|}{0.08}          & \multicolumn{1}{l|}{0.86}          & 0.87          \\ \hline
9                  & \multicolumn{1}{l|}{\textbf{85.76}} & \multicolumn{1}{l|}{\textbf{0.11}} & \multicolumn{1}{l|}{\textbf{12.12}} & \multicolumn{1}{l|}{\textbf{0.13}} & \multicolumn{1}{l|}{\textbf{0.73}} & \textbf{0.77} & \multicolumn{1}{l|}{\textbf{212.61}} & \multicolumn{1}{l|}{\textbf{0.18}} & \multicolumn{1}{l|}{\textbf{20.75}} & \multicolumn{1}{l|}{\textbf{0.07}} & \multicolumn{1}{l|}{\textbf{0.87}} & \textbf{0.88} \\ \hline
11                 & \multicolumn{1}{l|}{87.15}          & \multicolumn{1}{l|}{0.12}          & \multicolumn{1}{l|}{12.43}          & \multicolumn{1}{l|}{0.14}          & \multicolumn{1}{l|}{0.70}          & 0.74          & \multicolumn{1}{l|}{214.17}          & \multicolumn{1}{l|}{0.19}          & \multicolumn{1}{l|}{21.00}          & \multicolumn{1}{l|}{0.08}          & \multicolumn{1}{l|}{0.83}          & 0.85          \\ \toprule[2pt]
\end{tabular}
\label{tab:temperature}
\end{table}

\noindent
\paragraph{Effect of Temperature $t$ of Knowledge Distillation} To study the effect of the temperature $t$, we conduct a parameter study on both datasets, which is reported in Table~\ref{tab:temperature}. We can observe that the performance of \emph{LightPath} varies with different temperatures. It can be figured out that the best temperature $t$ is $9$, which indicates warm temperature can mitigate the peakiness of the teacher model and results in better performance. 

\begin{table}[!htp]
	
	\centering
	\caption{Effect of Balancing Factor $\alpha$}
	\begin{tabular}{l|llllll|llllll}
		\toprule[2pt]
		\multirow{3}{*}{\textbf{$\alpha$}} & \multicolumn{6}{l|}{\textbf{Aalborg}}                                                                                                                                                                                                  & \multicolumn{6}{l}{\textbf{Chengdu}}                                                                                                                                                                                                   \\ \cline{2-13} 
		& \multicolumn{3}{l|}{\textbf{Travel Time Estimation}}                                                           & \multicolumn{3}{l|}{\textbf{Path Ranking}}                                                                            & \multicolumn{3}{l|}{Travel Time Estimation}                                                                     & \multicolumn{3}{l}{Path Ranking}                                                                                     \\ \cline{2-13} 
		& \multicolumn{1}{l|}{\textbf{MAE}}   & \multicolumn{1}{l|}{\textbf{MARE}} & \multicolumn{1}{l|}{\textbf{MAPE}}  & \multicolumn{1}{l|}{\textbf{MAE}}  & \multicolumn{1}{l|}{\textbf{$\tau$}} & \textbf{$\rho$} & \multicolumn{1}{l|}{\textbf{MAE}}    & \multicolumn{1}{l|}{\textbf{MARE}} & \multicolumn{1}{l|}{\textbf{MAPE}}  & \multicolumn{1}{l|}{\textbf{MAE}}  & \multicolumn{1}{l|}{\textbf{$\tau$}} & \textbf{$\rho$} \\ \toprule[1pt]
		0                                               & \multicolumn{1}{l|}{90.24}          & \multicolumn{1}{l|}{0.12}          & \multicolumn{1}{l|}{12.78}          & \multicolumn{1}{l|}{0.16}          & \multicolumn{1}{l|}{0.69}                         & 0.73                         & \multicolumn{1}{l|}{220.32}          & \multicolumn{1}{l|}{0.19}          & \multicolumn{1}{l|}{21.52}          & \multicolumn{1}{l|}{0.09}          & \multicolumn{1}{l|}{0.79}                         & 0.81                         \\ \hline
		0.2                                             & \multicolumn{1}{l|}{89.35}          & \multicolumn{1}{l|}{0.12}          & \multicolumn{1}{l|}{12.85}          & \multicolumn{1}{l|}{0.14}          & \multicolumn{1}{l|}{0.69}                         & 0.73                         & \multicolumn{1}{l|}{217.10}          & \multicolumn{1}{l|}{0.19}          & \multicolumn{1}{l|}{21.34}          & \multicolumn{1}{l|}{0.09}          & \multicolumn{1}{l|}{0.78}                         & 0.80                         \\ \hline
		0.4                                             & \multicolumn{1}{l|}{91.57}          & \multicolumn{1}{l|}{0.12}          & \multicolumn{1}{l|}{13.17}          & \multicolumn{1}{l|}{0.15}          & \multicolumn{1}{l|}{0.69}                         & 0.73                         & \multicolumn{1}{l|}{217.33}          & \multicolumn{1}{l|}{0.19}          & \multicolumn{1}{l|}{21.21}          & \multicolumn{1}{l|}{0.08}          & \multicolumn{1}{l|}{0.85}                         & 0.87                         \\ \hline
		0.6                                             & \multicolumn{1}{l|}{\textbf{85.76}} & \multicolumn{1}{l|}{\textbf{0.11}} & \multicolumn{1}{l|}{\textbf{12.12}} & \multicolumn{1}{l|}{\textbf{0.13}} & \multicolumn{1}{l|}{\textbf{0.73}}                & \textbf{0.77}                & \multicolumn{1}{l|}{\textbf{212.61}} & \multicolumn{1}{l|}{\textbf{0.18}} & \multicolumn{1}{l|}{\textbf{20.75}} & \multicolumn{1}{l|}{\textbf{0.07}} & \multicolumn{1}{l|}{\textbf{0.87}}                & \textbf{0.88}                \\ \hline
		0.8                                             & \multicolumn{1}{l|}{87.44}          & \multicolumn{1}{l|}{0.12}          & \multicolumn{1}{l|}{12.76}          & \multicolumn{1}{l|}{0.14}          & \multicolumn{1}{l|}{0.70}                         & 0.75                         & \multicolumn{1}{l|}{214.34}          & \multicolumn{1}{l|}{0.19}          & \multicolumn{1}{l|}{20.93}          & \multicolumn{1}{l|}{0.08}          & \multicolumn{1}{l|}{0.84}                         & 0.86                         \\ \hline
		1                                               & \multicolumn{1}{l|}{89.23}          & \multicolumn{1}{l|}{0.12}          & \multicolumn{1}{l|}{12.78}          & \multicolumn{1}{l|}{0.16}          & \multicolumn{1}{l|}{0.69}                         & 0.73                         & \multicolumn{1}{l|}{215.03}          & \multicolumn{1}{l|}{0.19}          & \multicolumn{1}{l|}{21.02}          & \multicolumn{1}{l|}{0.08}          & \multicolumn{1}{l|}{0.82}                         & 0.84                         \\ \toprule[2pt]
	\end{tabular}
	\label{tab:glkdbf}
\end{table}
\begin{table}[htp]
	\centering
	\caption{Comparison with Whole Path Input}
	\begin{tabular}{l|lll}
		\toprule[2pt]
		\multirow{3}{*}{}            & \multicolumn{3}{l}{\textbf{Aalborg}}                                   \\ \cline{2-4} 
		& \multicolumn{3}{l}{\textbf{Travel Time Estimation}}                    \\ \cline{2-4} 
		& \multicolumn{1}{l|}{\textbf{MAE}}   & \multicolumn{1}{l|}{\textbf{MARE}} & \textbf{MAPE}  \\ \toprule[1pt]
		\emph{LightPath w/o RR} ($\gamma=0.1$) & \multicolumn{1}{l|}{91.97} & \multicolumn{1}{l|}{0.12} & 13.53 \\ \hline
		\emph{LightPath w/o RR} ($\gamma=0.1$) & \multicolumn{1}{l|}{90.85} & \multicolumn{1}{l|}{0.12} & 13.39 \\ \hline
		\emph{LightPath} ($\gamma=0.1$)        & \multicolumn{1}{l|}{82.79} & \multicolumn{1}{l|}{0.11} & 11.95 \\ \toprule[2pt]
	\end{tabular}
	\label{tab:fullpath}
	\vspace{-5pt}
\end{table}

\begin{table}[htp]
	\centering
	\caption{Comparison with Fixed Interval Strategy}
	\begin{tabular}{l|l|lll}
		\toprule[2pt]
		\multirow{3}{*}{\textbf{Strategy}}       & \multirow{3}{*}{$\gamma$} & \multicolumn{3}{l}{\textbf{Aalborg}}                                           \\ \cline{3-5} 
		&                           & \multicolumn{3}{l}{\textbf{Travel Time Estimation}}                            \\ \cline{3-5} 
		&                           & \multicolumn{1}{l|}{MAE}   & \multicolumn{1}{l|}{\textbf{MARE}} & \textbf{MAPE} \\ \toprule[1pt]
		\emph{LightPath-Fixed}  & 1/10                      & \multicolumn{1}{l|}{87.77} & \multicolumn{1}{l|}{0.12}          & 12.95         \\ \hline
		\emph{LightPath-Fixed}  & 5/10                      & \multicolumn{1}{l|}{89.63} & \multicolumn{1}{l|}{0.12}          & 12.86         \\ \hline
		\emph{LightPath-Fixed}  & 9/10                      & \multicolumn{1}{l|}{92.87} & \multicolumn{1}{l|}{0.12}          & 12.23         \\ \hline
		\emph{LightPath-Random} & 0.9                       & \multicolumn{1}{l|}{85.76} & \multicolumn{1}{l|}{0.11}          & 12.12         \\ \toprule[2pt]
	\end{tabular}
	\label{tab:fixstrategy}
\end{table}

\noindent
\paragraph{Effect of Balancing Factor $\alpha$} To study the effect of the balancing factor of global-local knowledge distillation, we conduct a parameter study on both datasets. Based on the results reported in Table~\ref{tab:glkdbf}, we observe that the performance of our model changes when varying $\alpha$. We can observe that the optimal $\alpha$ is 0.6, which means that global and local knowledge distillation loss can contribute to the \emph{LightPath}'s performance. When $\alpha=0$, the global knowledge distillation loss is ignored, which yields poor performance. When $\alpha=1.0$, the local knowledge distillation loss is ignored, and the performance also performs poorly. This confirms our conjecture that the
two proposed global-local knowledge distillation losses can regularize each other and
achieve better results than only optimizing one of them (i.e., $\alpha=0.0$ or $\alpha=1.0$).

\subsection{Comparison with Whole Path Input }
	
Compared with whole path input, we consider a variant “LightPath w/o RR” where only reconstruction loss is used and the relational reasoning (RR) loss is disabled. In this setting, we can see in the table \ref{tab:fullpath} that LightPath w/o RR ($\gamma=0$), i.e., using whole paths, outperforms LightPath w/o RR ($\gamma=0.1$), i.e., using partial paths. This means that, when only using the reconstruction loss, using whole paths are indeed better than using partial paths. However, when using both losses, LightPath ($\gamma=0.1$) outperforms LightPath w/o RR ($\gamma=0$). This demonstrates that the relational reasoning loss, which employs partial paths to create different views, is indeed effective.\emph{LigthPath}, 

\subsection{Comparison with Fixed Interval Strategy}
	
We then conducted additional experiments where we removed edges at fixed intervals strategy. Specifically, we set the fixed interval to 10 and removed $n$ edges out of every 10 (for instance, we deleted 1 edge out of every 10 edges, i.e., corresponding to a removal ratio of $\gamma = 0.1$). The results on travel time estimation in Aalborg shown in Table \ref{tab:fixstrategy}, which indicate that the fixed interval edge removal strategy (ref. as to first three rows) achieves the worse performance compared with the random edge removal strategy (ref. as to last row).

\subsubsection{Model Efficiency} 
\begin{figure}[t]

     \centering
     \begin{subfigure}[b]{0.45\textwidth}
         \centering
         \includegraphics[width=\textwidth]{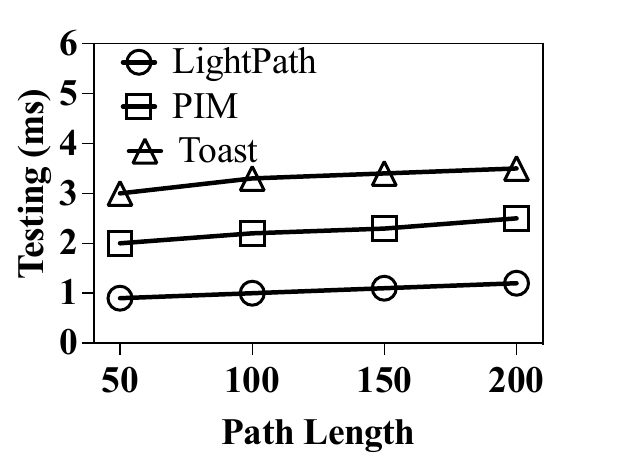}
         \caption{Training}
         \label{fig:subfig:training}
     \end{subfigure}
     \hfill
     \begin{subfigure}[b]{0.45\textwidth}
         \centering
         \includegraphics[width=\textwidth]{fig8b.pdf}
         \caption{Testing}
         \label{fig:subfig:inference}
     \end{subfigure}
        \caption{Model Efficiency Evaluation}
        \label{fig:mee}

\end{figure}

We finally evaluate the model efficiency, including training and inference phases. Figure~\ref{fig:mee} illustrates the corresponding results. The first observation is that \emph{LightPath} outperforms \emph{PIM} and \emph{Toast} in both training and inference phases. In the training phase, \emph{LightPath} is more than $3\times$ faster than \emph{PIM} and almost $5\times$ faster than \emph{Toast} when path length is 200. In the testing phase, we measure the running time for each path sample. As observed, \emph{LightPath} achieves up to at least 100\% and almost 200\% performance improvement compared with \emph{PIM} and \emph{Toast} when path length is 200.

\section{Related Work}
\subsection{Path Representation Learning}
Path Representation Learning (PRL) aims to learn effective and informative path representations in road network that can be applied to various downstream applications, i.e., travel cost estimation, and path ranking. Existing PRL studies can be categorised as supervised learning (SL) based~\cite{DBLP:conf/icde/Yang020,yang2020context,DBLP:journals/pvldb/0026HFZL020}, unsupervised learning (UL) based~\cite{DBLP:conf/ijcai/YangGHT021}, and weakly supervised learning (WSL) based~\cite{DBLP:journals/corr/abs-2203-16110} approaches. 
SL based methods aim at learning a task-specific path representation with the availability of large amounts of labeled training
data~\cite{DBLP:journals/pvldb/0026HFZL020,DBLP:conf/icde/Yang020,yang2020context}, which has a poor generality for other tasks. 
UL methods are to learn general path representation learning that does not need labeled training data and generalizes well to multiple downstream tasks~\cite{DBLP:conf/ijcai/YangGHT021,DBLP:journals/corr/abs-2203-16110}. In contrast, WSL methods try to learn a generic temporal path representation by introducing meaningful weak labels, e.g., traffic congestion indices, that are easy and
inexpensive to obtain, and are relevant to different tasks~\cite{DBLP:journals/corr/abs-2203-16110}. 
However, we aim to learn generic path representations instead of temporal path representations in this paper. Thus, we do not select WSL method as baseline method.
In particular, these methods are computationally expensive and hard to deploy in resource-limited environments.

\subsection{Self-supervised Learning}
State-of-the-art self-supervised learning can be classified into contrastive learning-based and relation reasoning-based methods. Contrastive learning-based methods~\cite{DBLP:journals/corr/abs-1807-03748,DBLP:journals/corr/abs-1808-06670,DBLP:conf/iclr/VelickovicFHLBH19,DBLP:conf/ijcai/YangGHT021,DBLP:conf/sigmod/BabaevOKIGNT22}, especially for InfoNCE loss-based, commonly generate different views of same input data through different augmentation strategies, and then discriminate positive and negative samples. However, these methods suffer from their quadratic complexity, w.r.t. the number of data samples, given that it needs a large number of negative samples to guarantee that the mutual information lower bound is tight enough~\cite{DBLP:journals/corr/abs-1808-06670}. 
In contrast, relation reasoning-based methods~\cite{DBLP:conf/nips/PatacchiolaS20,DBLP:journals/corr/abs-2011-13548} aim to learn relation reasoning
head that discriminates how entities relate to themselves and other entities, which results in linear complexity. However, existing studies construct relation reasoning between different views from the same encoder, ignoring the effect of different views between different encoders, i.e., main encoder and auxiliary encoder in Siamese encoder architecture.

\section{Conclusion}

We design a lightweight and scalable framework called \emph{LightPath} for unsupervised path representation learning. In this framework, we first propose sparse auto-encoder that is able to reduce path length $N$ to $N^{\prime}$, where $N$ is much larger than $N^{\prime}$, which in turn reduces the computation complexity of the model. Then, we use path reconstruction decoder to reconstruct the input path to ensure no edges information missing. Next, we propose a novel self-supervised relational reasoning approach, which contains cross-network relational reasoning and cross-view relational reasoning loss, to enable efficient unsupervised training. After that, we introduce global-local knowledge distillation to further reduce the size of sparse path encoder and improve the performance. Finally, extensive experiments on two real-world datasets verify the efficiency, scalability, and effectiveness of \emph{LightPath}.
%
%

\section*{Acknowledgments}
This work was supported in part by Independent Research Fund Denmark under agreements 8022-00246B and 8048-00038B, the VILLUM FONDEN under agreements 34328 and 40567.

\bibliographystyle{unsrt}  
\bibliography{references}

\end{document}